\documentclass[lettersize,journal]{IEEEtran}
\usepackage{amsmath,amsfonts}
\usepackage{algorithm}
\usepackage{array}
\usepackage[caption=false,font=normalsize,labelfont=sf,textfont=sf]{subfig}
\usepackage{textcomp}
\usepackage{stfloats}
\usepackage{url}
\usepackage{verbatim}
\usepackage{graphicx}
\usepackage{cite}
\usepackage{bbding}
\usepackage{tikz}
\usepackage{booktabs}
\usepackage{multirow}
\usepackage{algpseudocode}   
\usepackage{enumerate}
\usepackage{enumitem}
\usepackage{color}
\usepackage{algorithmicx,eqparbox} 
\usepackage{tabularx}
\usepackage{cite}
\usepackage{rotating}
\usepackage{lipsum}
\usepackage{nicematrix}
\usepackage{makecell} 

\begin{document}

\title{A Comparative Review of Recent Few-Shot Object Detection Algorithms}

\author{Jiaxu~Leng,~
        Taiyue~Chen,~
        Xinbo~Gao$^*$,~\IEEEmembership{Senior Member,~IEEE,}
        Yongtao~Yu, ~\IEEEmembership{Senior Member, ~IEEE,}
        Ye~Wang,~
        Feng~Gao,~
        and~Yue~Wang
\thanks{Jiaxu Leng, Taiyu Chen, Xinbo Gao, Ye Wang and Feng Gao are with the Key Laboratory of Image Cognition, Chongqing University of Posts and Telecommunications, Chongqing 400065, China}
\thanks{Yue Wang is with the School Of Cyber Security and Information Law, Chongqing University of Posts and Telecommunications, Chongqing 400065, China}
\thanks{Yongtao Yu is with the Faculty of Computer and Software Engineering, Huaiyin Institute of Technology, Huaian 223003, China}
\thanks{$^*$Corresponding author: Xinbo Gao.}}



\maketitle

\begin{abstract}

Few-shot object detection, learning to adapt to the novel classes with a few labeled data, is an imperative and long-lasting problem due to the inherent long-tail distribution of real-world data and the urgent demands to cut costs of data collection and annotation. Recently, some studies have explored how to use implicit cues in extra datasets without target-domain supervision to help few-shot detectors refine robust task notions.  This survey provides a comprehensive overview from current classic and latest achievements for few-shot object detection to future research expectations from manifold perspectives. In particular, we first propose a data-based taxonomy of the training data and the form of corresponding supervision which are accessed during the training stage. Following this taxonomy, we present a significant review of the formal definition, main challenges, benchmark datasets, evaluation metrics, and learning strategies. In addition, we present a detailed investigation of how to interplay the object detection methods to develop this issue systematically. Finally, we conclude with the current status of few-shot object detection, along with potential research directions for this field.

\end{abstract}


\section{Introduction}
\label{sec_introduction}
\IEEEPARstart{G}{iven} a set of classes, object detection aims to detect all instances of these classes in an/a image or video. As a fundamental task of computer vision, object detection has achieved great attention and been applied to numerous downstream applications, e.g., intelligent monitoring \cite{liu2019intelligent}, augmented reality \cite{liu2019edge}, automatic driving \cite{zhong2017class}.

\begin{figure}[!t]
\centering
\includegraphics[width=0.5\textwidth]{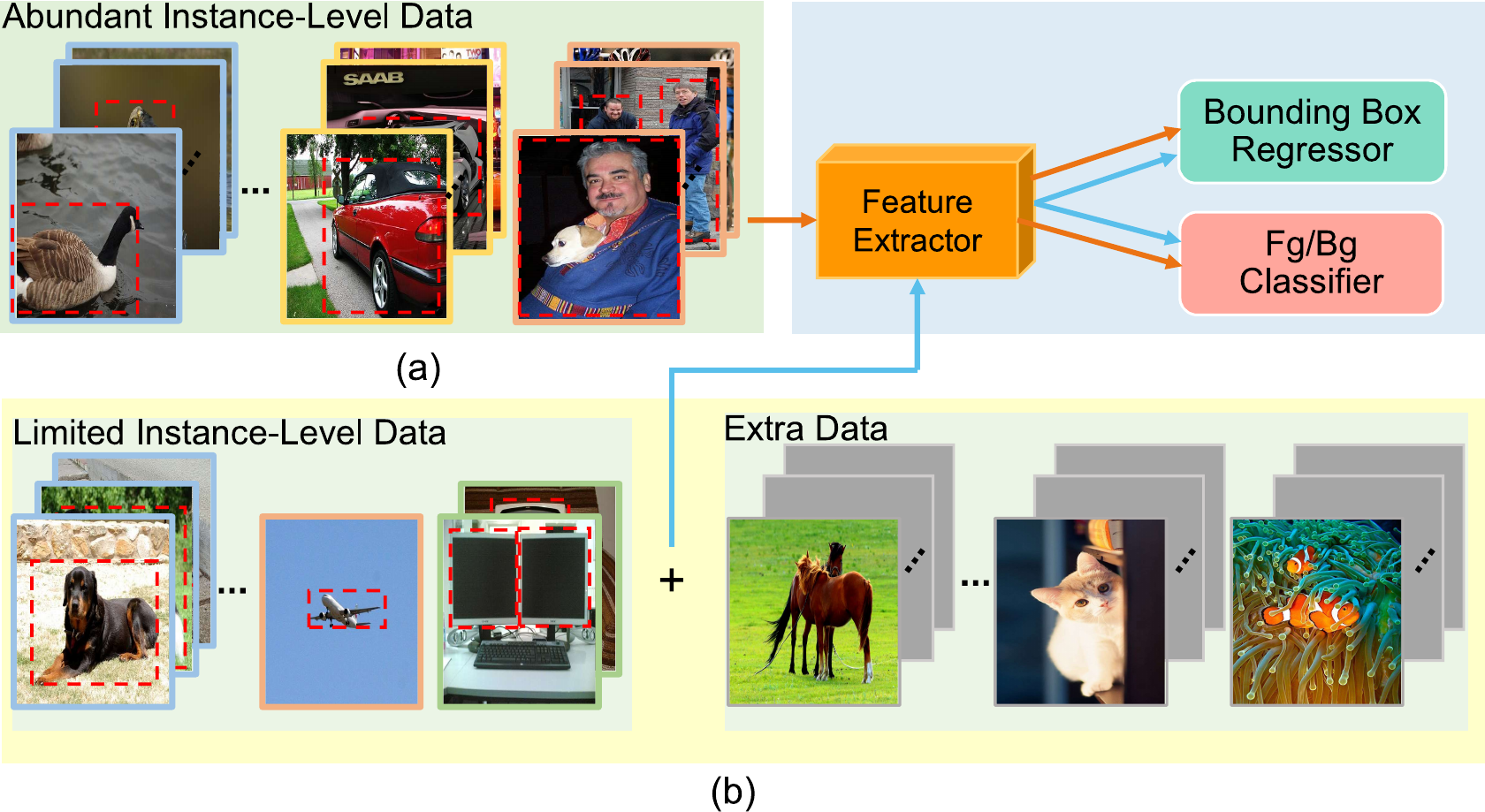}
\caption{Many-shot vs few-shot object detection. (a) The pipeline of many-shot object detection. It exploits a large-scale dataset with instance-level labels to learn a robust detector. (b) The pipeline of few-shot object detection. Only limited labeled data and extra datasets without target-domain supervision can be accessed during the training stage. Note that the formation of the target-domain supervision could be image-level tags. }
\label{fig_1}
\end{figure}

Earlier, traditional approaches attempted to exploit hand-crafted features to exhaustively search objects \cite{viola2001rapid,dalal2005histograms,felzenszwalb2010cascade,felzenszwalb2008discriminatively}, requiring abundant prior knowledge to manually design suitable features for special objects detection (e.g., face, pedestrian and traffic signs). Due to Alexnet’s remarkable performance on ImageNet in 2012 \cite{krizhevsky2012imagenet}, deep learning began to obtain increasing attention in
 the computer vision community, since it could automatically mine implicit task notions from training data and achieve huge performance gains when compared with traditional approaches. Especially in recent years, deep-learning approaches have made great breakthroughs in object detection \cite{ren2015faster,redmon2017yolo9000,liu2016ssd,redmon2018yolov3}. In order to extract robust concepts, deep learning models tend to acquire abundant labeled data for training. Nevertheless, it is not always easy to collect volumes of well-labeled data for a specific task: (1) data preparation is considerably time-consuming and laborious where it would cost about 10 seconds to label an instance \cite{zhang2021weakly}; (2) some rare cases exist at very low frequency, due to the inherent long-tail distribution of real-world data, e.g., endangered animals. Specifically, daily applications are crying out for few-shot learning to cut costs, while generic techniques and strategies could be prone to either capture noise as common notions (i.e., overfitting) or diverge (i.e., underfitting) in few-shot scenarios. However, even a child can quickly extract task-specific notions when shown small data and associated labels. Therefore, it encourages us to develop few-shot object detection (FSOD) that not only needs as little supervision as possible but also should be superior/close to many-shot detectors, as shown in Fig. \ref{fig_1}. Especially, we strictly restrict the total amount of supervision and do not limit the form of supervision. Here, we mainly discuss three main types of few-shot settings, as shown in Section \ref{taxonomy}.

In recent years, few-shot learning has achieved several crucial breakthroughs, especially in few-shot classification (FSC) \cite{koch2015siamese,vinyals2016matching,snell2017prototypical,oreshkin2018tadam,triantafillou2017few,chen2020new,finn2017model,nichol2018first,munkhdalai2017meta,munkhdalai2018rapid,bertinetto2016learning,gidaris2018dynamic,wang2019tafe,chen2018closer}. Inspired by recent advances in FSC, earlier works regarded FSOD as a FSC issue that first exploited a region proposal algorithm (e.g., SS \cite{uijlings2013selective}) to generate preliminary regions of interest (RoIs) and classify each RoI whether or not to contain an object. However, unlike FSC, FSOD is more complex, which not only needs to classify each RoI, but also should localize each RoI precisely. It is infeasible to look at two complementary sub-tasks in isolation. Earlier works have relatively low precision, since excessive low-quality and ambiguous proposals could confuse the meta classifiers. Subsequent works began to adopt a new scheme to simultaneously optimize few-shot detectors for both sub-tasks in order to obtain high-quality proposals. Especially, several metric-based approaches \cite{liu2020afd} provided category-specific notions (e.g., aspect ratios and sizes of objects) to the bounding box regressor. Moreover, existing approaches still rely on existing detectors, e.g., R-CNN, YOLO and SSD variants \cite{girshick2014rich, ren2015faster, redmon2017yolo9000, redmon2018yolov3,bochkovskiy2020yolov4,liu2016ssd}, which were originally designed to tackle many-shot issues, and do not take special considerations into few-shot issues. Classic architectures not only should exhaustively search all locations whether to cover objects or not, but also need to associate features with object shapes, which also requires that backbones should effectively and efficiently encode both shape and class notions into semantics for objects of novel classes. However, in low-shot scenarios, too large and too low intra-class variations are very common where large intra-class variations tend to bring with low inter-class distinction and low intra-class variations usually lead to low data diversity (e.g., aspect ratios). It is hard to exploit limited data to learn a robust encoder and thus few-shot detectors cannot extract high-quality proposals from non-robust features. Therefore, many FSOD approaches utilized extra datasets \cite{deng2009imagenet,chen2015microsoft} to acquire generic notions (e.g., pre-trained backbones \cite{he2016deep,xie2017aggregated,simonyan2014very,krizhevsky2012imagenet}) for these heavy-weight frameworks, which were conducive to tackle few-shot challenges, instead of training from scratch. To obtain high performance, several works supposed that a novel category has close relation with base categories, e.g., shared visual components (color/shape/texture), adding extra constraints (i.e., KL divergence) to efficiently transfer shared notions to novel classes. However, it led to some new issues, e.g., domain shifts \cite{wang2019few,wu2020multi}, where source-domain knowledge would not fit target domain well. In that case, such pre-training phases could have little effect for a novel task and FSOD approaches could very easily confuse highly similar classes and have uncertainty in localizing objects of novel classes \cite{wu2020multi,chen2018lstd,wang2020frustratingly,kang2019few}, due to little inter-domain and noisy intra-domain support (Section \ref{subsec_challenges}). Moreover, most FSOD methods focus on a classic $N$-way $K$-shot setting since it needs not consider an imbalanced problem and 
has no requirement to obtain implicit information from extra unlabeled data collected from the target domain when compared with other classic settings, in Section \ref{taxonomy}. In brief, FSOD still has a long way to go.

Here, we limit the scope of this paper on how to learn a competent detector under few-shot/limited-supervised settings. For content completeness, we also present a compendious review of advances in object detection, few-shot learning, semi-supervised learning and weakly-supervised learning. The main contributions are summarized as follows: 
\begin{itemize}{}{}
\item{We identify few-shot problems and propose a novel data-based taxonomy for studying main challenges and existing solutions in FSOD.} 
\item{We summarize existing solutions in a systematic manner. The outline of our survey includes the definition of the few-shot problems, benchmark datasets, evaluation metrics, a summary of the main approaches. Specially, for these approaches, we provide a detailed analysis of how these methods interplay with each other to promote development in this promising field. }
\item{We present and discuss the potential research directions in this issue. }
\end{itemize}

The overall organization is presented in Fig. \ref{overview}. We first provide a brief review for recent advances in related tasks, such as few-shot learning, in Section \ref{sec_background}. Following the taxonomy proposed in Section \ref{sec_introduction}, we respectively present their definition, benchmark datasets, evaluation metrics and related works of how to learn robust few-shot detectors under various limited-supervised settings in Section \ref{sec_limited}, Section \ref{sec_semi} and Section \ref{sec_weakly}. Specially, in Section \ref{sec_limited}, we also involve how they work together to promote technical progress. Finally, we conclude this survey, along with the potential future trends in this promising domain, in Section \ref{sec_conclusion}.

\subsection{Comparison with Previous Reviews}
In recent years, miscellaneous generic object detection surveys have been published \cite{zhao2019object,sharma2017review,dhillon2020convolutional,zou2019object,zhiqiang2017review,liu2020deep,sultana2020review}. Zhao et al. \cite{zhao2019object}, Sharma et al. \cite{sharma2017review} and Dhillon et al. \cite{dhillon2020convolutional} provided a detailed analysis, including classic architectures, useful tricks, benchmarks, evaluation metrics, etc. Wang et al. \cite{zhiqiang2017review}, Liu et al. \cite{liu2020deep} and Sultana et al. \cite{sultana2020review} highlighted recent developments of deep-neural-network based detectors. Especially, there exist numerous excellent surveys that focused on a kind of detectors designed for several specific objects, such as pedestrian detection \cite{li2012review,hurney2015review,li2006review}, moving object detection \cite{kulchandani2015moving,hatwar2018review,karasulu2010review}, face detection \cite{kumar2019face, nivedharecent}, traffic sign detection \cite{liu2019machine,deshpande2016brief,mukhometzianov2017machine} and so on. Oksuz et al. \cite{oksuz2020imbalance} and Chen et al. \cite{chen2020foreground} presented imbalance problems existing in deep-neural-network based detectors, e.g., foreground-background imbalance. Unlike 
previous surveys, we focus on the few-shot challenge for object detection which does not systematically appear in previous surveys.

As aforesaid, few-shot learning has received great attention in the computer vision community and many well-written works have summarized it profoundly \cite{wang2020generalizing, kadam2018review, li2020concise, jadon2020overview, rezaei2020zero, yin2020meta, li2021deep}. Wang et al. \cite{wang2020generalizing} and Kadam et al. \cite{kadam2018review} indicated core issues in few-shot learning and grouped few-shot approaches into three categories (i.e., data-, model- and algorithm-based methods). Li et al. \cite{li2020concise} provided a comprehensive analysis of meta-learning methods. Jadon \cite{jadon2020overview} discussed deep-neural-network architectures designed for few-shot learning. Apart from these surveys for generic few-shot learning, several surveys also put emphasis on specific applications of few-shot learning: COVID-19 diagnosis \cite{rezaei2020zero}, natural language processing \cite{yin2020meta} and computer vision \cite{li2021deep}. Here, except classic $N$-way $K$-shot settings, we also discuss imbalance problems, semi-supervised settings and weakly-supervised settings for FSOD, while previous surveys failed to cover these concepts.

In addition, FSOD aims at learning to simultaneously localize and classify all instances from a few labeled data, which is more challenging and under-explored. We notice that existing approaches are still fragmentary and unsystematic and there is no relative survey to present the current development status and tendency in FSOD, which is detrimental to carry out solid researches. Consequently, it is essential to present a comprehensive survey of related works on FSOD and reveal inner relations and motivations about how they promote development of this promising task. We hope that our thorough survey can provide insights for further research.

\subsection{Taxonomy}
\label{taxonomy}
Although abundant excellent FSOD works \cite{chen2018lstd,wang2020frustratingly,kang2019few} have been published recently, they are proposed under different settings or for separate objectives in terms of data settings, training strategies and network architectures, where it is improper to discuss them together. Due to limited supervision, most few-shot detectors must rely on extra datasets to give an appropriate initialization for these generic yet heavy-weight frameworks, while there were large discrepancies among their settings of extra datasets. Thus, according to data and associated supervision which could be accessed during the training stage, as shown in Tab. \ref{tab_fsod_categories}, we group these approaches into three categories - limited-supervised based FSOD (LS-FSOD), semi-supervised based FSOD (SS-FSOD) and weakly-supervised based FSOD (WS-FSOD), respectively. In LS-FSOD, there is only a small dataset $D_{novel}$ with a few instance-level labeled exemplars of each novel class to learn novel task notions and an optional dataset without target supervision to learn generic notions. Unlike LS-FSOD, there is an extra target-domain dataset  $D_{novel}^-$ without annotations in SS-FSOD to enforce few-shot detectors to automatically capture implicit objects for reducing labour force. In WS-FSOD, there is a small dataset $D_{novel}^+$ with a few image-level labeled exemplars of each novel class to enforce few-shot detectors to mine implicit relations among image-level tags and associated objects. In some cases, we also include a target-domain dataset $D_{novel}^-$ and a base dataset $D_{base}$ to compensate inaccurate supervisory signals from $D_{novel}^+$ to make the training process more stable.

\begin{table}[!t]
  \centering
  \caption{Comparison of three main types of FSOD. $\surd$/$\times$/$\bigcirc$ indicates inclusive/exclusive/optional, respectively.} 
    \centering
    \begin{tabular}{|c|c|c|c|c|}
    \hline
    Type & $D_{novel}$ & $D_{novel}^+$ & $D_{novel}^-$ & $D_{base}$ \\
    \hline
    LS-FSOD & $\surd$ & $\times$ & $\times$ & $\bigcirc$ \\
    \hline
    SS-FSOD & $\surd$ & $\times$ & $\surd$ & $\bigcirc$ \\
    \hline
    WS-FSOD & $\times$ & $\surd$ & $\bigcirc$ & $\bigcirc$ \\
    \hline
    \end{tabular}
    \label{tab_fsod_categories}
\end{table}

\begin{figure*}[!t]
\centering
\includegraphics[width=\textwidth]{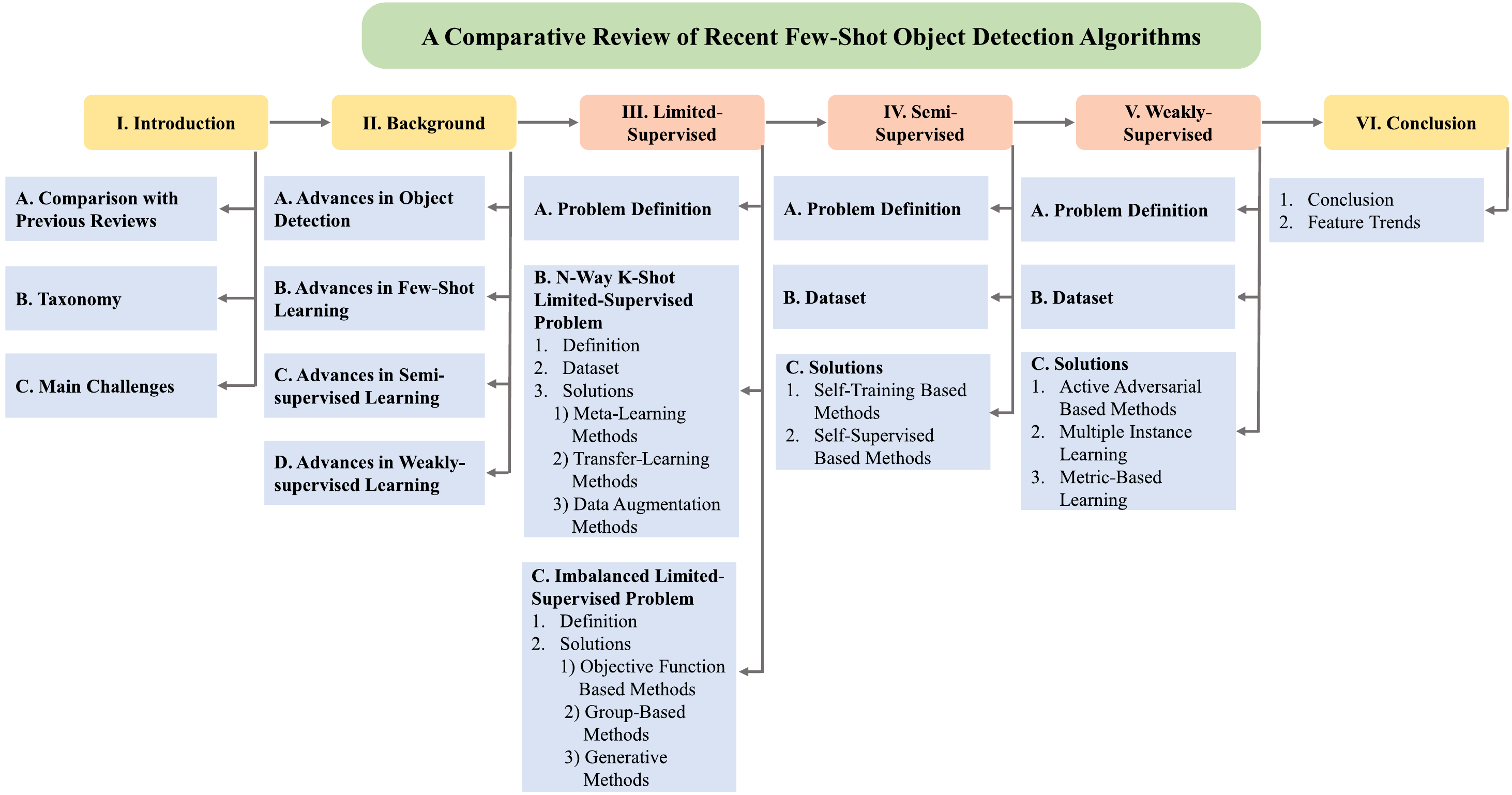}
\caption{Structure of this survey.}
\label{overview}
\end{figure*}

In all, it is infeasible to only rely on limited labeled data to learn a robust model. As aforesaid, even a child could perform so well with very little training data and it raises a question of why humans could quickly adapt to a new task. Dubey et al. \cite{dubey2018investigating} proved that humans can exploit history/prior knowledge to tackle their confronted issues. Referring to the human learning process, the key is how to extract task-agnostic notions from $D_{base}$ or capture task-specific guidance from $D_{novel}^+$ or $D_{novel}^-$, which is helpful for a novel few-shot task. However, due to supervision differences, there exist huge technical gaps among three types of settings. For example, compared with LS-FSOD, SS-FSOD could exploit a large-scale unlabeled dataset that contains objects of novel classes during training, which means SS-FSOD must introduce a special framework to mine underlying objects or generic notions from unlabeled data (Section \ref{sec_semi}). In the following sections, 
we will elaborate the specific definition, challenges and existing approaches respectively for all kinds of FSOD. In conclusion, according to above requirements, we summarize a set of tools for FSOD, as shown in Fig. \ref{overview}.

\subsection{Main Challenges} 
\label{subsec_challenges}
Currently, few-shot detectors still adopt classic deep-learning frameworks, e.g., R-CNN, YOLO and SSD variants \cite{girshick2014rich,bochkovskiy2020yolov4,liu2016ssd}, which are inevitable to confront with intrinsic challenges, e.g., imbalance problems \cite{li2020overcoming,ge2021delving,phan2020resolving}, large intra-category variations \cite{zhang2012implicit} and low inter-category distance (fine-grain problems) \cite{angelova2013efficient,lv2021fine}. Besides, limited supervisory signals could further make some issues more serious, where low-density sampling has a high probability to lead to nasty data distributions, e.g., high intra-class variations, low inter-class distance and data shift. Thus, few-shot detectors still need to develop suitable learning strategies to overcome the degradation phenomenon that deep-learning approaches are prone to acquire irrelevant features (i.e., overfitting), which cannot be corrected automatically due to inadequate support. Here, we will discuss main issues that make the training process more challenging:

\begin{figure}[!t]
\centering
\includegraphics[width=0.5\textwidth]{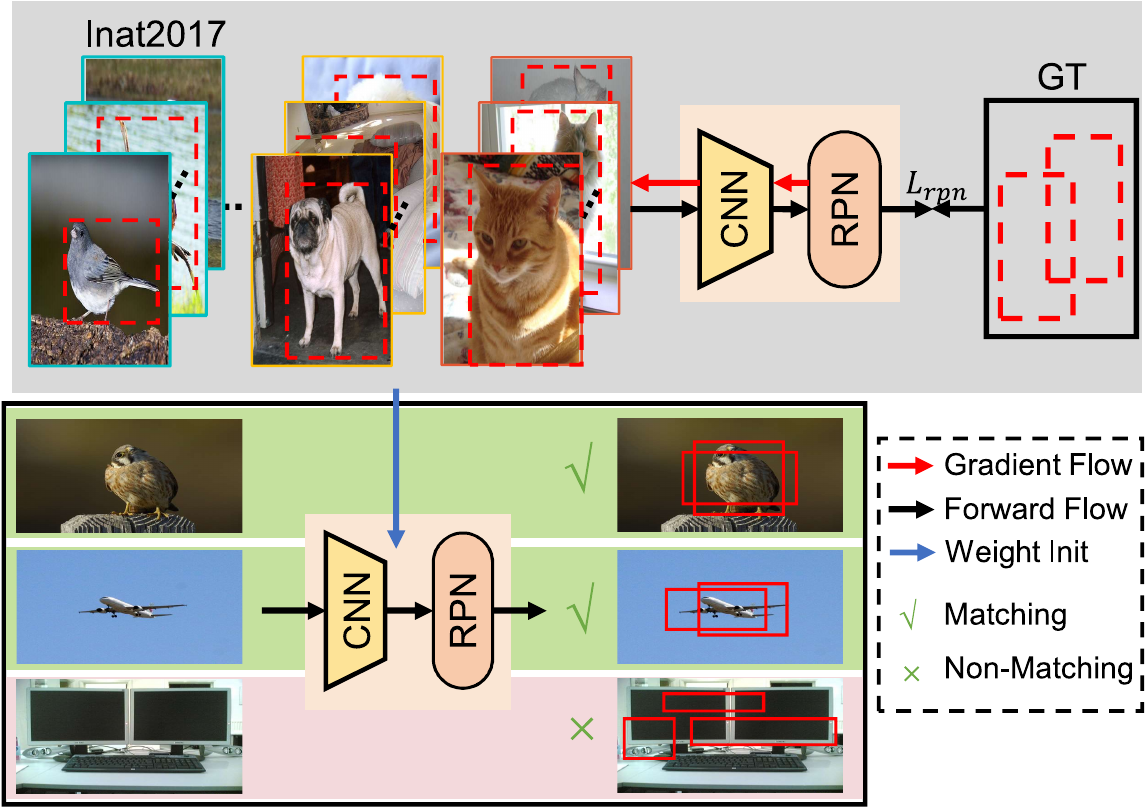}
\caption{Performance of RPN with large/small domain shift. The backbone and RPN are pretrained on the inat2017 dataset \cite{van2018inaturalist} (only animals \& plants). Bird is a class of the inat2017 dataset while the RPN generates good proposals. Although aeroplane is not a class of the inat2017 dataset, aeroplane shares similar visual components with bird and gets comparable results. TV/monitor has large domain differences with the inat2017 dataset and gets worse results.}
\label{fig_domain_shifts}
\end{figure}

\begin{figure}[!t]
\centering
\includegraphics[width=0.5\textwidth]{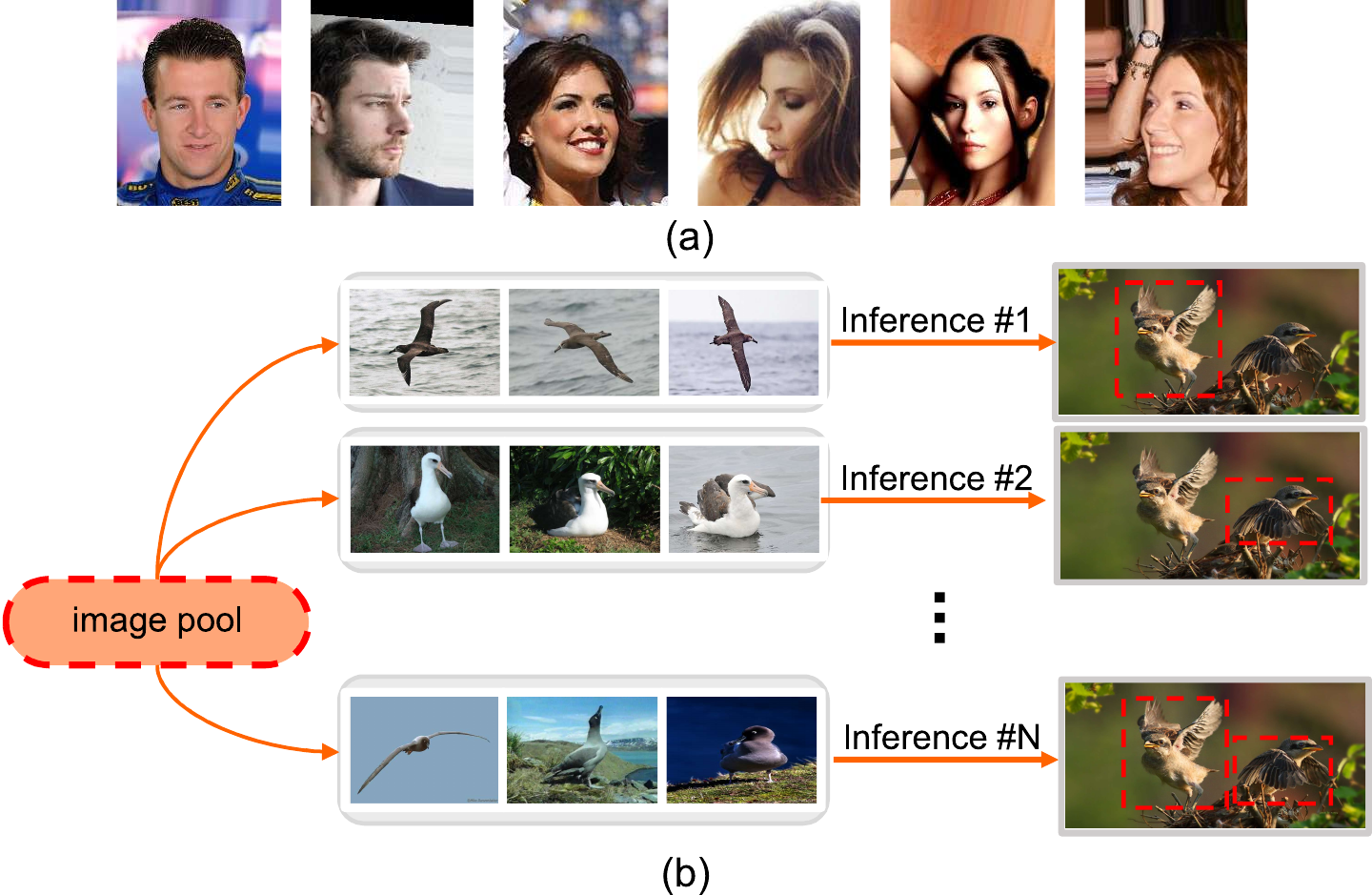}
\caption{ (a) Several exemplars of data bias. There exist noisy backgrounds and pose variations among these training images. (b) The negative effect of data bias. The quality of training images sampled from the image pool could seriously influence final performance.}
\label{fig_data_bias}
\end{figure}

\begin{itemize}{}{}
\item{\textbf{Domain Shifts.} Numerous FSOD approaches utilized a large-scale dataset to learn generic notions that are subsequently transferred/fine-tuned to meet requirements for a novel task. In some cases, the source domain shares little cross-domain knowledge with the target domain, i.e., large domain shifts, where “generic” notions learned from the source domain produce weak or even negative effect for the target task. For instance, many works hold a principle that the region proposal network (RPN) was an ideal proposal algorithm which could generate high-quality regions of interest (RoIs) for all foreground classes \cite{liu2020afd, bansal2018zero,yang2020restoring,li2021beyond}. However, foreground classes, which are defined as a set of classes of interest, are task-specific, while all other classes are defined as negatives. Therefore, such a class-agnostic RPN cannot provide RoIs for novel classes as good as that for base classes, especially with large domain shift \cite{fan2020few, wu2020meta}. As illustrated in Fig. \ref{fig_domain_shifts}, the similarity between the base and novel classes has a great effect on the quality of RoIs for novel classes. In addition, the low quality of RoIs (e.g., TV/monitor) will degrade the subsequent training of detection heads. Thus, generic notions should be used carefully when existing domain shift.} 
\item{\textbf{Data Bias.} A dataset is essentially a collection of observed exemplars from a special data distribution. In reality, there are large intra-class variations even for the same objects, such as appearance, posture and so on, while large intra-category variations could relatively fuzzy decision boundaries, as illustrated in Fig. \ref{fig_data_bias}. Unlike large-scale datasets, it is impossible to cover all situations for a small-scale dataset that naturally has more data bias in scale, context, intra-class diversity and so on \cite{wu2020multi}. Due to large capacities for deep-learning methods, they could be susceptible to noise/bias to utilize non-robust notions to make decisions (i.e., overfitting). Especially, metric-learning based methods need to leverage the training set to learn a set of robust category prototypes as task-specific parameters. It is hard to build robust class prototypes when the training set has many outliers, such as occlusion. 

Due to low-density sampling, there may exist domain shifts among the training dataset and the testing dataset and various dataset splitting may produce unstable results. For a relatively reliable result, the average performance from multiple runs is a feasible way.
}

\item{\textbf{Insufficient Instance Samples.} Insufficient supervision could amplify implicit noise and data bias in the dataset, which could easily lead traditional and deep-learning approaches to overfitting or even underfitting. Especially, insufficient supervision tends to form a loose cluster for each class and it is unfeasible to learn a robust detector via increasing the intra-category diversity, such as abundant simple and naive data augmentation methods. }

\item{\textbf{Inaccurate Supervisory Signals.} In WS-FSOD, this issue is mainly caused by the ambiguous relation among image-level tags and associated objects. Due to inaccurate supervision, it has some difficulty associating each image-level tags with the whole objects appeared in this image and measuring the quality of these proposals, where it tends to associate these image-level tags with the most discriminative parts of these objects in most cases. In addition, insufficient image-level tags could further have a negative effect on this process, where it may have more inaccurate proposals. }

\item{\textbf{Incomplete Annotation.} As aforesaid, few-shot detectors usually need extra datasets to learn robust notions to initialize these heavy-weight learning frameworks. Due to annotation discrepancies, there may exist objects of novel categories in the base dataset $D_{base}$ and it is time-consuming and laborious to relabel these objects. When few-shot detectors are pre-trained  on this base dataset, they could treat these objects of novel categories as negatives and learn to suppress these objects, which is harmful to detect novel-class objects. Actually, this issue could be viewed as a semi-supervised problem. Thus, Li et al. \cite{li2021few} exploited a semi-supervised solution to mine these background proposals which probably contained objects of novel categories and assign pseudo labels to these proposals to train a novel detector. Due to lack of supervision, pseudo boxes could contain too many background regions and propagate noise into the subsequent training process.
}
\end{itemize}

\section{Background}
\label{sec_background}
In this section, we propose a brief review of advances in object detection, few-shot learning, semi-supervised learning and weakly-supervised learning. To present a better understanding of this paper, it is essential for readers to have some background knowledge of these topics. However, if you want to acquire detailed background knowledge, we suggest several latest and comprehensive surveys for readers \cite{wang2020generalizing,zou2019object,ouali2020overview,zhou2018brief}.

\subsection{Advances in Object Detection}

\begin{figure}[!t]
\centering
\includegraphics[width=0.5\textwidth]{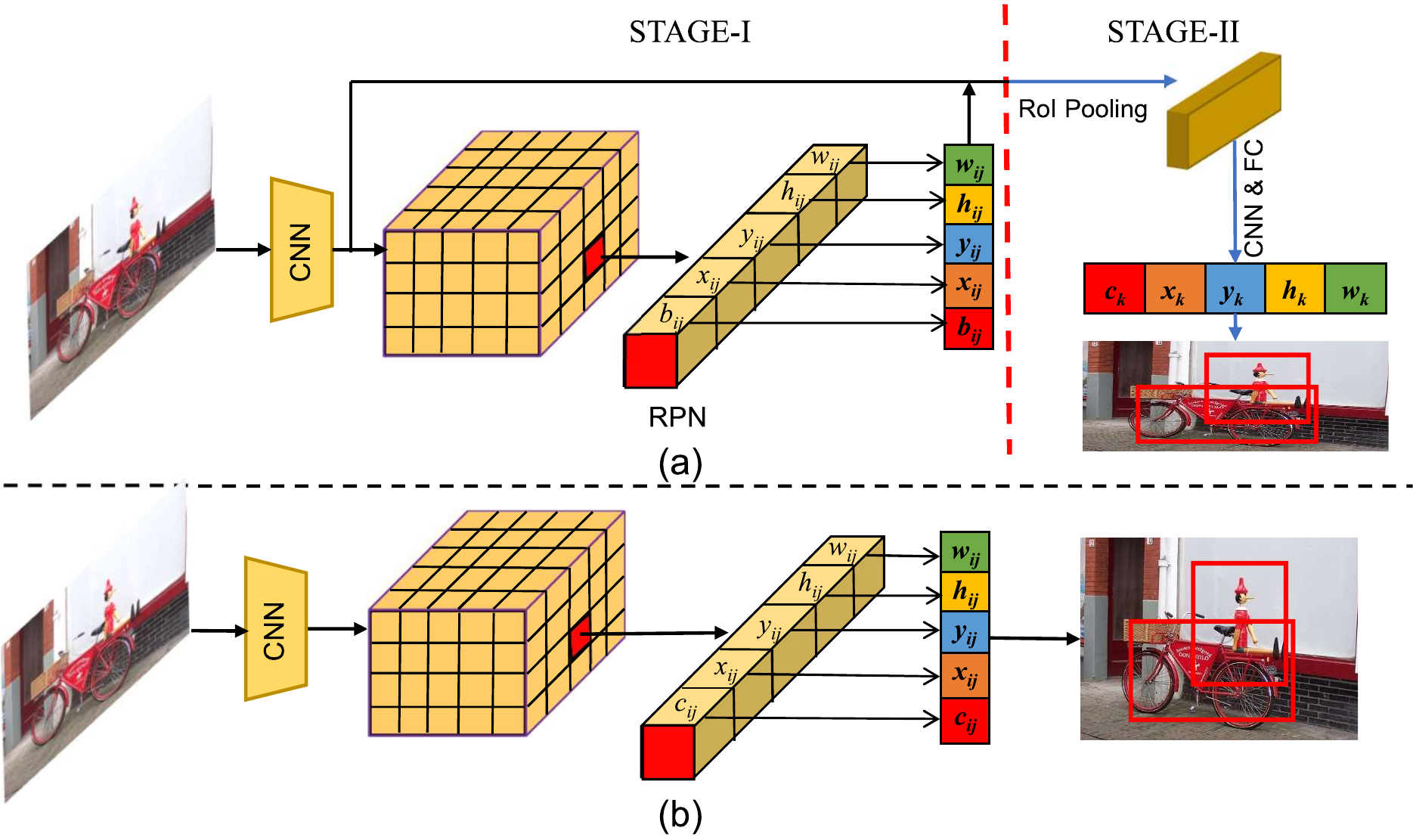}
\caption{(a) The pipeline of a classic two-stage object detection method (Faster R-CNN \cite{ren2015faster}), (b) The framework of a classic single-stage object detection method (YOLO \cite{redmon2016you}).}
\label{fig_obj}
\end{figure}

As aforementioned, object detection focuses on simultaneously localizing and classifying all objects of classes of interest. Currently, mainstream detection frameworks could be usually categorized into two types, i.e., one- and two-stage detectors. The main difference between one- and two-stage detectors is that two-stage detectors (e.g., R-CNN \cite{girshick2014rich}) follow the coarse-to-fine mechanism (i.e., the coarse and fine localization) while one-stage detectors (e.g., YOLO \cite{redmon2016you}) directly make final results without the extra refinement process (Fig. \ref{fig_obj}). Two-stage detectors first take a region proposal algorithm (e.g., RPN \cite{ren2015faster} and selective search (SS) \cite{uijlings2013selective}) to screen preliminary class-agnostic candidates (the coarse localization), then extract fixed-length vectors for all candidates by ROI-Align/Pooling, and finally feed them into two parallel branches and post-processing algorithms (e.g., non-maximum suppress (NMS) \cite{uijlings2013selective}) to produce final proposals (the fine location) (Fig. \ref{fig_obj}(a)). One-stage detectors, like SSD/YOLO/CornerNet-style algorithms \cite{liu2016ssd, redmon2016you, law2018cornernet, duan2019centernet}, directly generate a set of bounding boxes and associated category probability distributions for each spatial location, which are similar to the class-specific RPN (Fig. \ref{fig_obj}(b)).

Likewise, these classic frameworks could be grouped into anchor-based and anchor-free detectors as well, according to whether to exploit prior anchors during the process of proposal generation. Prior anchors are a set of pre-defined boxes with various aspect ratios and sizes and are initially used to provide fairly good reference for RPN to avoid too large search space. However, it is non-intuitive to specify hyper-parameters (i.e., the aspect-ratios, sizes and number) of anchors. Although several works (e.g., dimension clusters \cite{redmon2017yolo9000}, GA-RPN \cite{wang2019region}) have been proposed to tackle such problems, we cannot completely drop all hyper-parameters of prior anchors. In 2018, Law, H. and Deng, J. \cite{law2018cornernet} re-introduced the anchor-free mechanism and viewed it as a task of keypoint detection and matching. So far, keypoint has many kinds of definitions, such as left-top and right-bottom corners of objects \cite{law2018cornernet}, four extreme points (top-most, left-most, bottom-most, right-most) and one center point of objects \cite{zhou2019bottom}, etc.

Except for generic approaches above, there exist many excellent approaches designed to tackle special problems, such as the imbalance problem \cite{lin2017focal,shrivastava2016training,cai2018cascade,pang2019libra}, the real-time problem \cite{redmon2017yolo9000,redmon2016you,liu2016ssd}, small target detection \cite{li2017perceptual,kisantal2019augmentation,eggert2017closer,ozge2019power}, fine-grained object detection \cite{zhang2014part,angelova2013efficient, lv2021fine,song2020fine}, few-shot learning \cite{chen2018lstd, liu2020afd,fan2020few} and so on. These works have further promoted applications of object detection in real scenes.

\subsection{Advances in Few-Shot Learning}
Few-shot learning aims to learn to enhance the generalization ability with limited labeled data. In the deep-learning era, solutions of few-shot learning can be classified into three main types: meta-learning, transfer-learning and data augmentation methods. Meta-learning methods usually use abundant episodes/tasks to acquire task-agnostic notions (e.g., meta-parameters), which can be meaningful to quickly adapt to a new task. In addition, meta-learning methods can be further divided into three types, i.e., metric-, optimization- and model-based methods. Metric-based methods pay attention to learn a robust embedding function and a scoring function that measures similarity between embedding vectors of a query image and each class prototype\cite{koch2015siamese, vinyals2016matching, sung2018learning, snell2017prototypical,oreshkin2018tadam}. Sung et al. \cite{sung2018learning} proposed an end-to-end automatic scoring module to measure similarity. Koch et al. \cite{koch2015siamese} proposed a siamese network to obtain representatives for both query and support images, then utilized L1 distance to fuse these representatives, and finally fed the fused features to a MLP for evaluating similarity. In addition, metric-learning was also formed as a task of an information retrieval \cite{oreshkin2018tadam}. Optimization-based methods attempted to learn a meta-optimizer or meta-parameters for quick adaption to a new task \cite{finn2017model}. MAML variants \cite{finn2017model} took a two-step strategy to learn meta-parameters for a given task. Meta agents (e.g., LSTM) were designed to learn updating rules, such as learning rate. Model-based methods 
design a specific network architecture and corresponding learning strategies for quick adaption for a novel task \cite{gidaris2018dynamic,wang2019tafe,graves2014neural,santoro2016meta,munkhdalai2017meta}. Several works \cite{wang2019tafe,munkhdalai2017meta} grouped weights into two types (i.e., task-specific and task-agnostic parameters) and only update task-specific weights for quick and robust adaption to a new task. NTMs \cite{graves2014neural} and MANN \cite{munkhdalai2017meta} introduced LSTMs and cache pools for quickly generating task-specific parameters. Transfer-learning methods mainly rely on fine-tuning general notions from source datasets to a novel task without training from scratch. Chen et al. \cite{chen2018closer} proved that simple transfer-learning methods were much effective, even with large domain shifts.

Furthermore, several works explored data augmentation for increasing data diversity to mitigate overfitting. A GAN variant was proposed to transfer intra-class variations from base classes to novel classes for more robust prototypes \cite{wang2018low,hariharan2017low}.

\subsection{Advances in semi-supervised learning}
To get rid of over-dependence on abundant labeled data for deep-learning based methods, a large number of semi-supervised methods explore a new paradigm to enforce a learner to automatically acquire instance-level notions from partially annotated data \cite{lee2013pseudo,sohn2020simple,zhou2021instant}. Clearly, there exists hidden target-domain knowledge in unlabeled data and the key is to accurately extract this knowledge to regularize detectors. In machine learning, a classic way was to learn a teacher model from annotated data first, then apply it to generate pseudo labels as ground-truth labels for unlabeled data, and finally sample a list of reliable pseudo labels to train a student model \cite{zhou2005semi,valizadegan2008semi}. The quality of pseudo labels played a key role in the training process of the student learner. Therefore, multiple teacher models were ensembled to work together to produce stable pseudo labels \cite{jiang2012semi}. Besides, several works proposed graph-based methods which used all labeled and unlabeled data and their mutual relations to build a graph for label propagation to obtain pseudo labels for unlabeled data \cite{zhu2005semi,subramanya2014graph}.

In addition, several works took it as a clustering problem and hold a core hypothesis that all exemplars should have high inter-class distinction and low intra-class variations \cite{grandvalet2005semi,bennett1999semi}. Thus, S3VM variants are designed to find a proper decision boundary which ought to pass through low data-density regions \cite{bennett1999semi}.

Recently, deep-learning approaches have received large attention in the field of semi-supervised learning due to their large capacity \cite{lee2013pseudo,sohn2020simple,zhou2021instant,jeong2019consistency,rasmus2015semi}. Similarly, it is sub-optimal to simply use labeled data and drop underlying notions of unlabeled data in training phase. CNNs were equipped to mine pseudo labels in unlabeled data to train a student learner. 
Except for pseudo labels, unlabeled data is also exploited to learn a pretext task (e.g., a reconstruction task) which enforces a learner to keep consistent among multiple models or recover raw signals from features for a better encoder, e.g., $\pi$-model, temporal ensembling and mean teacher \cite{jeong2019consistency,rasmus2015semi,odena2016semi,laine2016temporal}.

\subsection{Advances in weakly-supervised learning}
To lower instance-level annotation burdens, weakly-supervised object detection (WSOD) attempts to exploit a relative cheap alternative (e.g., image-level tags or object locations) and automatically mine underlying cues among weak supervision and objects for a novel task, which has received much attention recently \cite{yang2019towards,tang2018pcl,bilen2014weakly,cheng2020high}.

However, unlike instance-level boxes, these cheap alternatives cannot be directly applied to guide models to localize objects and instead bring uncertain supervisory signals into the training process. In weakly-supervised learning, it tends to propagate image-level tags for estimating instance-level boxes as ground-truth pseudo labels for weakly annotated data. Thus, weakly-supervised approaches have to tackle uncertainty problems raised by imprecise supervision, where detectors do not have a suitable way to measure the quality of pseudo boxes. Uncertainty problems tend to conclude with two aspects, i.e., low-quality pseudo boxes and inaccurate labels. Generally, detectors tend to associate image-level tags with the most discriminative parts of objects which leads to too small pseudo boxes \cite{bilen2016weakly,zhou2016learning}. Likewise, if multiple instances are clustered in unlabeled images, detectors could even attempt to utilize a large box to cover all instances which brings too large pseudo boxes. Moreover, when negatives have a certain IoU with foreground objects, it is hard for detectors to suppress background proposals for high-quality pseudo boxes and thus generate inaccurate labels. Thus, it still has huge performance gaps between state-of-the-art weakly-supervised and fully-supervised approaches so far \cite{shamsolmoali2021multipatch,wang2021end}. 

For high-quality pseudo labels, it could be grouped into two kinds, i.e., initialization \cite{pandey2011scene,wang2013weakly} and refinement \cite{bilen2016weakly,wang2021end,tang2017multiple,zhou2016learning}. Initialization aims to distinguish suitable pseudo boxes from abundant proposals, using prior knowledge. There exist many kinds of prior knowledge, including saliency/foreground heatmaps, inter-category variations, intra-category similarity, object co-occurrence and so on. As for refinement, it mainly focuses on learning strategies which could alleviate negative effects from inaccurate pseudo boxes, e.g., multiple instance learning (MIL) \cite{bilen2016weakly,wang2021end,tang2017multiple} and class activation map (CAM) \cite{zhou2016learning,zhang2018adversarial}. MIL-style approaches aim to mine underlying evidence among image-level tags and all RoIs in a specific image, i.e., class probability distributions for each RoI and object distributions for each class, where two kinds of evidence could be aggregated into class probability distributions for such an image. It could guide models to automatically distinguish various kinds of instances under a setting with reliable supervision, instead of directly assigning a pseudo label for each box. CAM-style approaches attempt to use image-level labels to learn a classifier first and slide its weights upon image features to generate class-sensitive heatmaps which could be segmented as bounding boxes. Except these works, there exist a lot of works that restricted their scope into some specific classes, e.g., pedestrians \cite{yu2016weakly, htike2014weakly, cai2016pedestrian}, vehicles \cite{jiang2016weakly,cao2017weakly,chadwick2020radar}, face \cite{huang2017learning} and so on. 

\section{Limited-Supervised Few-Shot Object Detection}
\label{sec_limited}
LS-FSOD is a classic issue in the few-shot learning which only relies on very limited supervision to learn task notions for novel classes. To alleviate overfitting, it usually exploits large-scale open-access datasets \cite{deng2009imagenet,van2018inaturalist,chen2015microsoft,everingham2010pascal} to mine task-agnostic notions which are helpful to any other tasks. To achieve stable support from the base dataset, it requires that the source task should share some generic notions with the novel task (Section \ref{subsec_challenges}). In this section, LS-FSOD can be grouped into two types: balanced and imbalanced LS-FSOD, according to whether there exists the foreground-foreground imbalance problem in the novel dataset. Here, we have mainly discussed solutions of the former and propose candidate solutions for the latter, although the latter hasn't been raised great attention.

\subsection{Problem Definition}
As shown in Tab. \ref{tab_fsod_categories}, let $C_{base}$ be a set of classes in a large-scale dataset $D_{base}$. Similarly, $C_{novel}$ is a set of classes in a small-scale dataset $D_{novel}$ with instance-level labels. Here, we assume $D_{base}$ with instance-level labels for simplicity. Note that $C_{base}$ and $C_{novel}$ are disjointed, i.e., $C_{base} \cap C_{novel} = \emptyset$. For each sample $(I, Y)$ in $D_{base}\cup D_{novel}$, $I$ is an image ($I \in \mathbb{R}^{M \times N \times 3}$) and $Y=\{(b_n, y_n)\}^N$ is a list of $N$ objects in $I$, where $b_n \in \mathbb{R}^4$ is the bounding box of the n-th instance and $y_n \in \{0, 1\}^{|C_{base} \cup C_{novel}|}$ denotes an associated one-hot label encoding. Especially, we mainly evaluate the performance of the novel categories $C_{novel}$.

Let $D_{novel}^n$ be a set of all objects of the n-th class in $D_{novel}$. For $D_{novel}$, we restrict the maximum of the number of instances per class in $D_{novel}$: $\max{\{|D_{novel}^i|,i\in C_{novel}\}} \le k$ ($k$ is usually no more than 30). Generally speaking, LS-FSOD attempts to acquire generic notions to mitigate too large parameter search space with very limited supervision from $D_{novel}$.

\subsection{$N$-Way $K$-Shot Limited-Supervised Problem}
\subsubsection{Definition} 
In the $N$-way $K$-shot setting, $N$ denotes the number of categories in $D_{novel}$ (i.e., $|C_{novel}|$) and $K$ is the number of objects per category (i.e., $\max{\{|D_{novel}^i|,i\in C_{novel}\}}=\min{\{|D_{novel}^i|,i\in C_{novel}\}}=k$).

\subsubsection{Dataset}
For convenience, benchmarks of the $N$-way $K$-shot problem consist of two sub-benchmarks (i.e., the base and novel dataset) which are usually built upon existing generic OD benchmarks, e.g., PASCAL VOC 07/12 \cite{everingham2010pascal}, MSCOCO \cite{chen2015microsoft} and ImageNet-LOC \cite{deng2009imagenet}. For a detailed review on these generic object detection datasets, we refer readers to the latest and comprehensive surveys \cite{zaidi2021survey, jiao2021new}. Here, we list common settings of benchmarks in $N$-way $K$-shot FSOD in Tab. \ref{tab_benchmarks}. Especially, in PASCAL VOC 07+12 \cite{everingham2010pascal}, the splitting settings have been a standard configuration in FSOD, i.e., \{(bird, bus, cow, motorbike, sofa / rest), (aero, bottle, cow, horse, sofa / rest), (boat, cat, motorbike, sheep, sofa / rest)\} \cite{liu2020afd,yan2019meta,wang2020frustratingly,wu2020multi}. Similarly, the splitting setting of FSOD and MSCOCO has been publicly available, as shown in Tab. \ref{tab_benchmarks}. However, the novel/base class splittings of an existing dataset may cause that the base dataset has many objects of novel classes, which is similar to the incomplete annotation problem. To tackle this problem, a simple way is to remove all images with objects of novel classes in the base dataset $D_{base}$ or view all instances of novel categories as background \cite{liu2020afd}. To make full use of latent exemplars in the base dataset $D_{base}$, Li et al. \cite{li2021few} allocated pseudo labels for these negative proposals which not only have low IoU with all ground-truth boxes but also have high similarity with novel category prototypes to learn a more reliable detector in meta-training stage (a semi-supervised style solution). Classic metrics in MSCOCO and $mAP_{50}$ are generally exploited to evaluate the actual performance of FSOD methods.

\begin{table*}[!t]
  \centering
  \caption{Common settings of benchmarks in $N$-way $K$-shot FSOD. Note that $B$/$N$ represents base/novel categories. Meanwhile, we also give a set of frequently-used values of $K$.}
    \centering
    \begin{tabular}{|c|c|}
    \hline
    Benchmarks & Settings \\
    \hline   
    PASCAL VOC 07+12 & 5B \& 5N ($K \in \{1,2,3,5,10\}$) \cite{liu2020afd,yan2019meta,wang2020frustratingly,wu2020multi}\\
    \hline    
    MSCOCO & 60B \& 20N (overlapped with PASCAL VOC 07+12) ($K \in \{10,30\}$) \cite{liu2020afd,yan2019meta,wang2020frustratingly,wu2020multi} \\
    \hline    
    FSOD  & 300/500/800B \& 200N ($K \in \{5\}$) \cite{fan2020few}\\
    \hline    
    MSCOCO $\rightarrow$ PASCAL VOC 07+12 & 60B (MSCOCO) \& 20N (PASCAL VOC 07+12) ($K \in \{10\}$) \cite{liu2020afd,yan2019meta,wu2020multi}\\
    \hline    
    MSCOCO $\rightarrow$ ImageNet-Loc & 80B (MSCOCO) \& 50N (ImageNet-Loc) ($K \in \{5\}$) \cite{fan2020few} \\
    \hline    
    MSCOCO $\rightarrow$ FSOD & 80B (MSCOCO) \& 200N (FSOD) ($K \in \{5\}$) \cite{fan2020few} \\
    \hline
    \end{tabular}
    \label{tab_benchmarks}
\end{table*}

\subsubsection{Solutions}
There are three types of solutions for the $N$-way $K$-shot problem: (a) meta-learning methods \cite{liu2020afd,fan2020few,hsieh2019one,yin2020meta}, (b) transfer-learning methods \cite{yang2020context,chen2018lstd,wang2020frustratingly}, (c) data augmentation methods \cite{zhang2021hallucination}. As aforesaid, meta-learning (learn-to-learn) puts more emphasis on exploiting $D_{base}$ to elaborately design a set of tasks to learn task-agnostic notions for quick adaption to a new task when compared with transfer-learning. Data augmentation methods highlight how to increase the intra-category diversity.

\emph{1. Meta-Learning Methods.} In the meta-learning, a task can be formulated as $\mathcal{T}_i={(\mathcal{S}^1,\mathcal{S}^2,\ldots,\mathcal{S}^N,Q)}$ where $\mathcal{S}^j$ denotes the j-th support images and Q is a query image. First, we exploit $D_{base}$ to construct a meta-training task set $\mathcal{T}^{train}$ for acquiring generic notions. Then, a fine-tuning task set $\mathcal{T}^{finetuning}$ is adapted to fit in the new task space, which is unnecessary 
for metric-based methods. Finally, a meta-testing task set $\mathcal{T}^{test}$ is applied to evaluate the adapted model. Especially, meta-learning methods can be further grouped into metric-, model- and optimization-based methods, according to the way to acquire task-agnostic notions in the meta-training stage.

\begin{figure}[!t]
\centering
\includegraphics[width=0.5\textwidth]{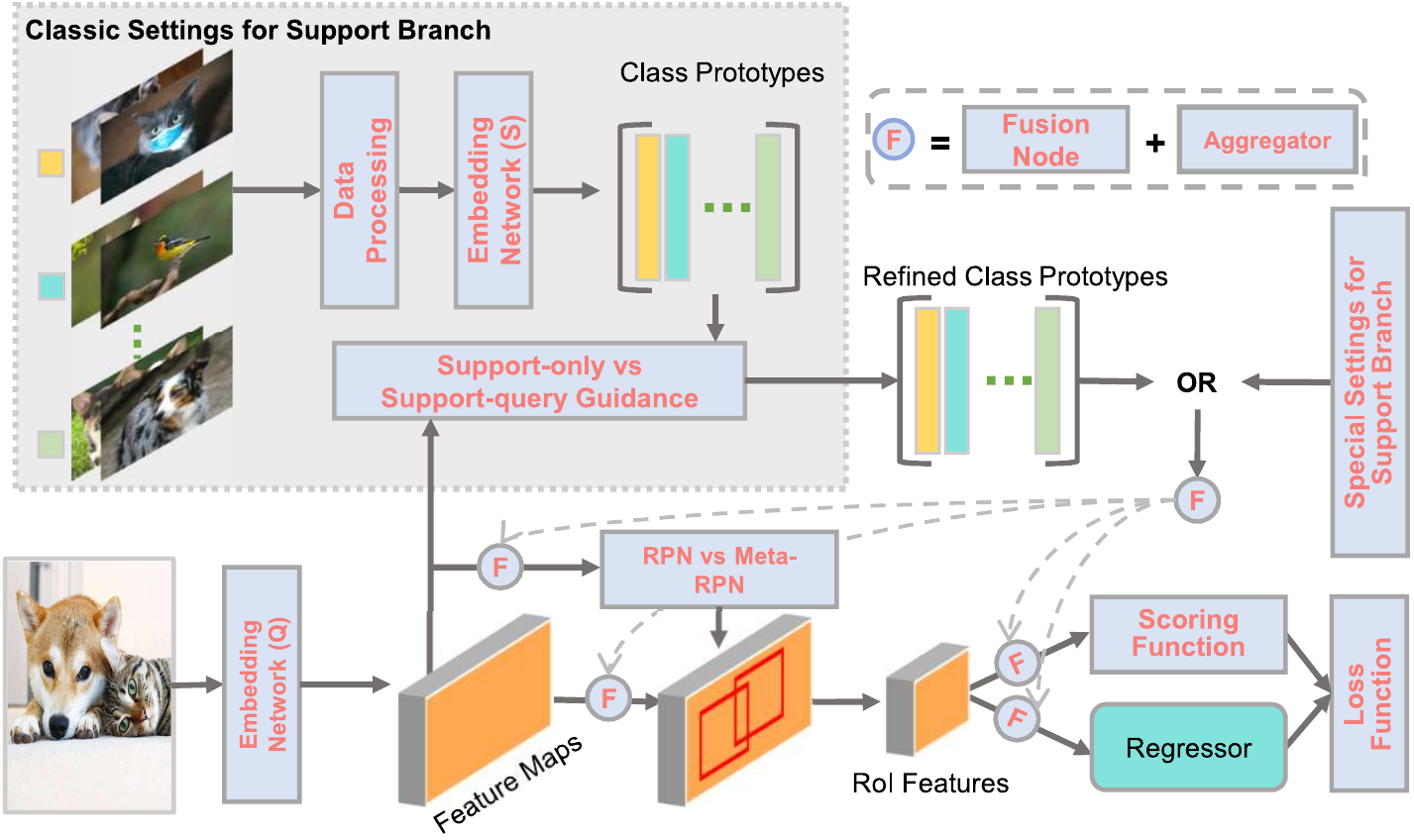}
\caption{Structure of Two-Stage Metric-Based Methods. \textcircled{F} represents a fusion node and an feature aggregator. Note that we only show several fusion nodes for simplicity and more details about fusion nodes are in Fig. \ref{fig_fusion_nodes} and Tab. \ref{tab_fusion_node}.}
\label{fig_metric_learning}
\end{figure}

\textbf{Metric-based Methods.} In this section, we will make a detailed review on metric-based detectors from ten aspects, i.e., the data processing, the embedding network, RPN vs meta-RPN, support-only vs support-query guidance, the aggregator, the scoring function, the loss function, the fusion node, the training/testing process and other settings for the support branch, as shown in Fig. \ref{fig_metric_learning}. These methods have two data flows, i.e., the support and query branch. The support branch takes responsibility for providing task-specific parameters and the query branch is in charge of combining task-specific parameters and query features to generate proposals. These branches exchange information via various fusion nodes and aggregators. To tackle the few-shot issue, existing methods have proposed customed solutions in each aspect, while they are fragmentary and unsystematic. Thus, we group their works into these aspects and give a brief review for each approach in Tab. \ref{metric_learning_all_methods_review} as well, if you want to know the complete strategy for a specific method. In addition, we also provide a detailed analysis of how they interactive to have huge performance gains, as shown in Fig. \ref{fig_metric_learning}.
\begin{itemize}
\item{{\bf{Data Preprocessing.}} In the support branch, all support instances for a specific category are usually extracted from support images by ground-truth boxes to generate a list of fixed-length category representatives for this category (Fig. \ref{fig_data_preprocessing}). However, this approach ignores contextual information (e.g., the co-occurrence of objects), which can be employed to exploit inter-category relations to get better class representatives. For more contextual information, Fan et al. \cite{fan2020few} and Han et al. \cite{han2021meta} directly added 16-pixel image context to each support instance. Wu et al. \cite{wu2020meta} extracted instance representatives from feature maps to implicitly use contextual information outside instances to enhance class representatives. Another way is to leverage ground-truth boxes to generate a binary foreground mask $M\in\mathbb{R}^{H\times W}$ that tends to be stacked with the corresponding image along the channel axis (i.e., $[I, M]$) to guide the network not only to focus on object area but also to acquire contextual information \cite{liu2020afd,li2021beyond,li2021transformation,yan2019meta}. To learn a more robust detector, Li et al. \cite{li2021beyond} proposed a feature disturbance method to augment $M$, which truncated 15\% pixels in $M$ to zero. Here, 15\% pixels were chosen as these with larger gradient, which tended to correspond the most discriminative features of objects, to enforce it to explore other equally good features (like dropout). In addition, the shape information of objects is essential to build appropriate representatives for a special category, e.g., the aspect ratios. To keep the structure of instances, there exist two main solutions: (1) prior to resize cropped patches to a fixed size, zero-padding is adopted to adjust aspect ratios of cropped patches \cite{chen2021should}; (2) a RoI-Pooling/Align layer is applied to get fixed-length features \cite{zhang2021meta,fan2020few}.}

\begin{figure}[!t]
\centering
\includegraphics[width=0.5\textwidth]{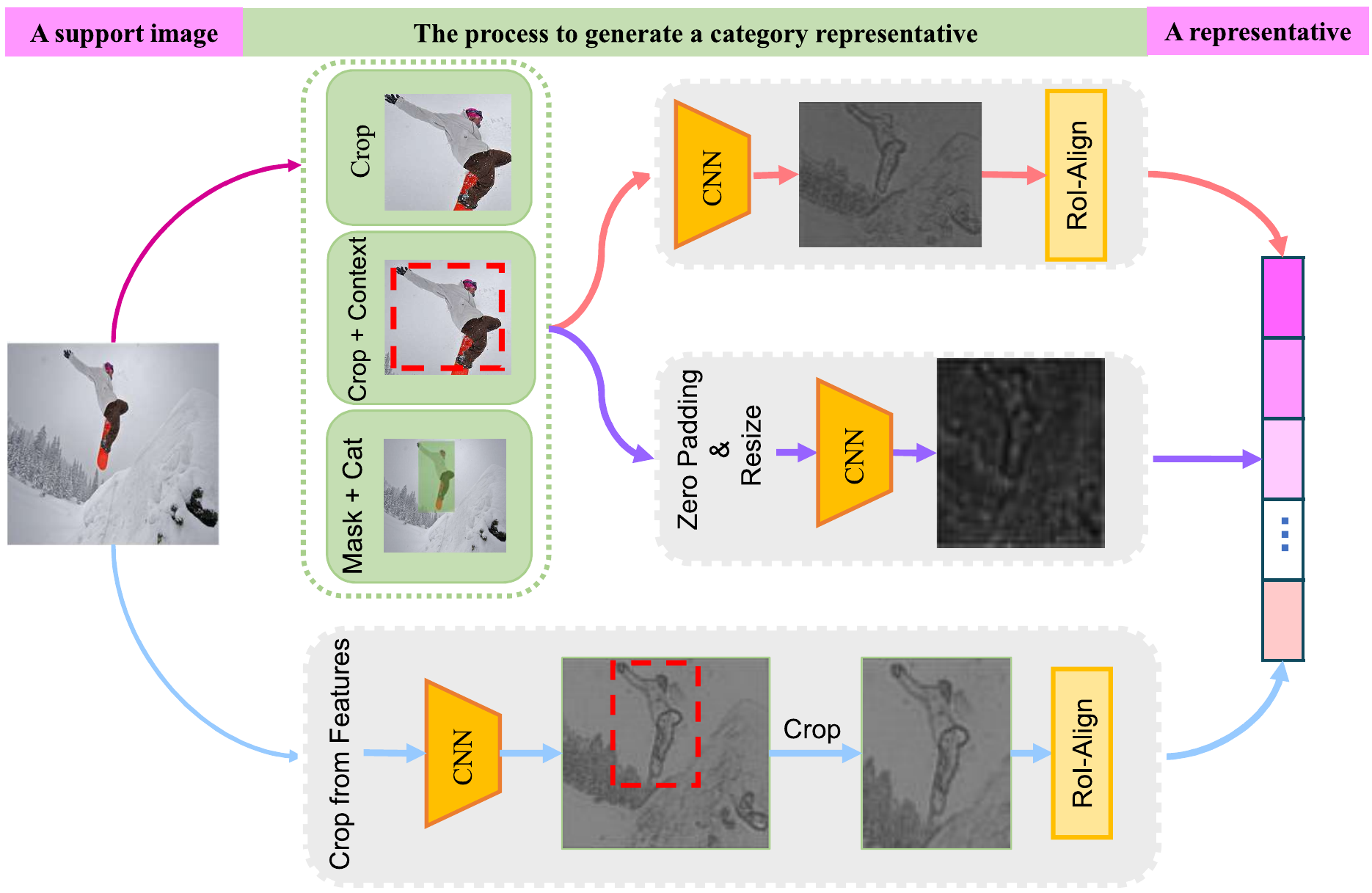}
\caption{The process to generate a fixed-length category representative.}
\label{fig_data_preprocessing}
\end{figure}

\item{\textbf{Embedding Network.} An embedding network (backbone) is pretty important in metric learning \cite{vinyals2016matching, koch2015siamese}. It consists of parallel sub-networks $f$ and $g$ (usually $f=g$) respectively for the query and support set. For the support set, all instances of a specific category are fed into $g$ to generate a list of instance representatives, and a clustering algorithm (e.g., k-means/median) is then applied to generate category prototypes from these instance representatives of a given category \cite{li2020one, yang2020restoring}. Likewise, a query image is fed into $f$ to generate corresponding query features for matching. 

In the metric learning, it relies on an important hypothesis that object features in a query image should have high similarity with prototypes of the corresponding category. However, due to low-shot scenarios, it is likely to have noise in the support set, such as occlusion, which is harmful to generate category prototypes. To solve such a problem, Li et al. \cite{li2021transformation} proposed a TIP to impose a consistency constraint on the features of an image and a corresponding corrupted image obtained by applying transformations on the image to make $f$/$g$ invariant to transformations. A similarity-based sampling strategy was designed to pair a query image with the most similar support instances in the training stage to alleviate intra-class variations \cite{li2020one} while it could use biased class prototypes during the inference stage. Similarly, scale differences among objects of the query and support set could violate the semantic consistency to confuse the scoring network. Singh et al. \cite{singh2018sniper} proved that there was a significant semantic difference between features of a given image at various scales, even if the siamese network was used to compute features. FPN was employed to match features of the support and query set at each scale to reduce scale differences. Nevertheless, FPN introduced more negative proposals while positive proposals were limited. Namely, it enlarged the foreground-background imbalance. To get more positive samples, Wu et al. \cite{wu2020multi} designed object pyramids to provide more positive supervision for each level of FPN. Zhang et al. \cite{zhang2020few} presented a multi-scale fusion module, which adopted up-sampling (i.e., bilinear interpolation) and down-sampling (i.e., 1x1 strided convolution) methods to map all features to the same scale, to explicitly mix scale information into feature maps. Meanwhile, unlike FPN, it greatly reduced negative proposals.}

\item{\textbf{RPN vs Meta-RPN.} In RPN, it was regarded as a category-agnostic algorithm which took a foreground-background classifier to screen RoIs regardless of their actual category. Earlier, most works supposed that a pretrained RPN could generate high-quality proposals for a novel task, and tended to freeze all parameters of such an RPN to avoid overfitting \cite{liu2020afd,bansal2018zero,yang2020restoring,li2020mm,zhu2021semantic}. In the pretraining stage, only a base dataset with limited classes available was utilized to learn such a class-agnostic RPN which could not cover all situations to produce the same good proposals for unseen classes, especially with large domain shift. As aforesaid, the notion of foreground classes is task-specific. To tackle this problem, traditional RPN was revised as meta-RPN which took query features conditioned by a set of prototypes for each category as input and only output associated RoIs of that category (Fig. \ref{fig_rpn}(b)) \cite{han2021meta,zhang2021accurate,li2020one}. Meta-RPN relied on a hypothesis that a set of prototypes of a given category could provide class-specific notions for RPN not only to suppress heterogeneous/background features but also to enhance similar semantics for more high-quality proposals. Zhang et al. \cite{zhang2020cooperating} proposed CoRPNs to ensemble multiple independent yet cooperative RPN to improve the quality of proposals (Fig. \ref{fig_rpn}(c)).}

\begin{figure}[!t]
\centering
\includegraphics[width=0.5\textwidth]{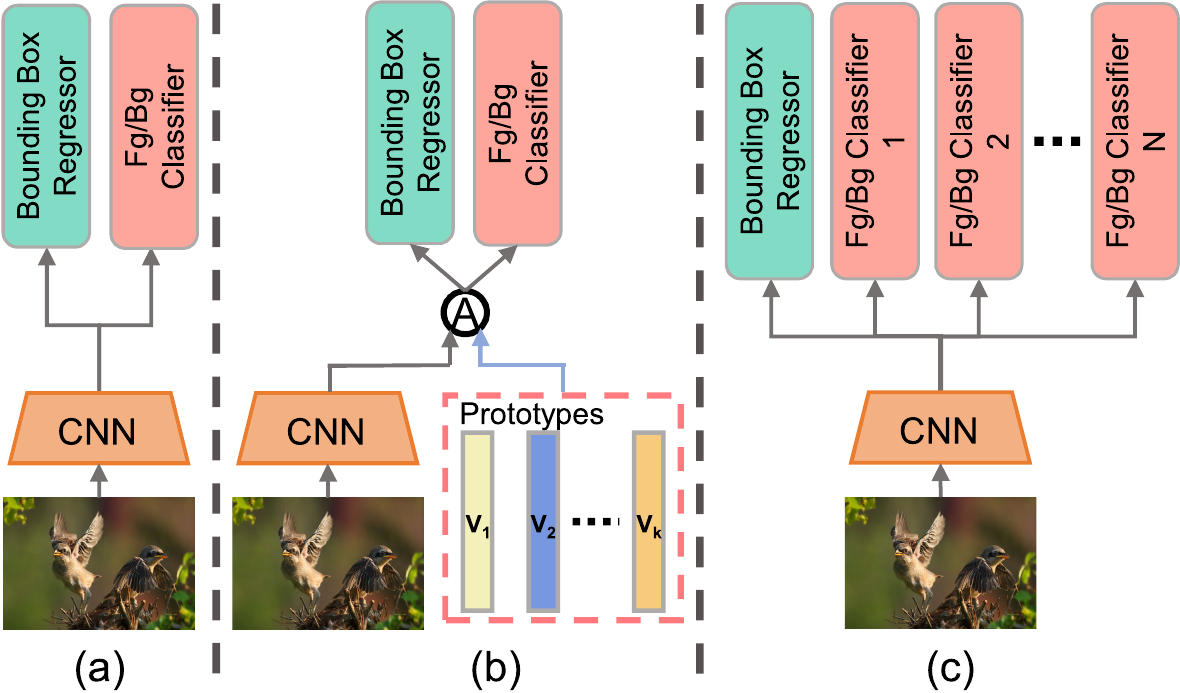}
\caption{(a) The pipeline of original RPN; (b) The architecture of meta-RPN. Let \textcircled{A} be a feature aggregator which takes query features and a set of representatives of a given category (i.e., $v_c={v_1,\ v_2,\ldots,\ v_k}$) as input and outputs category-specific query features that are fed to subsequent RPN; (c) The architecture of CoRPNs. Unlike original RPN, it consists of multiple Fg/Bg classifiers which is independent yet cooperative.}
\label{fig_rpn}
\end{figure}

\begin{figure*}[!t]
\centering
\includegraphics[width=\textwidth]{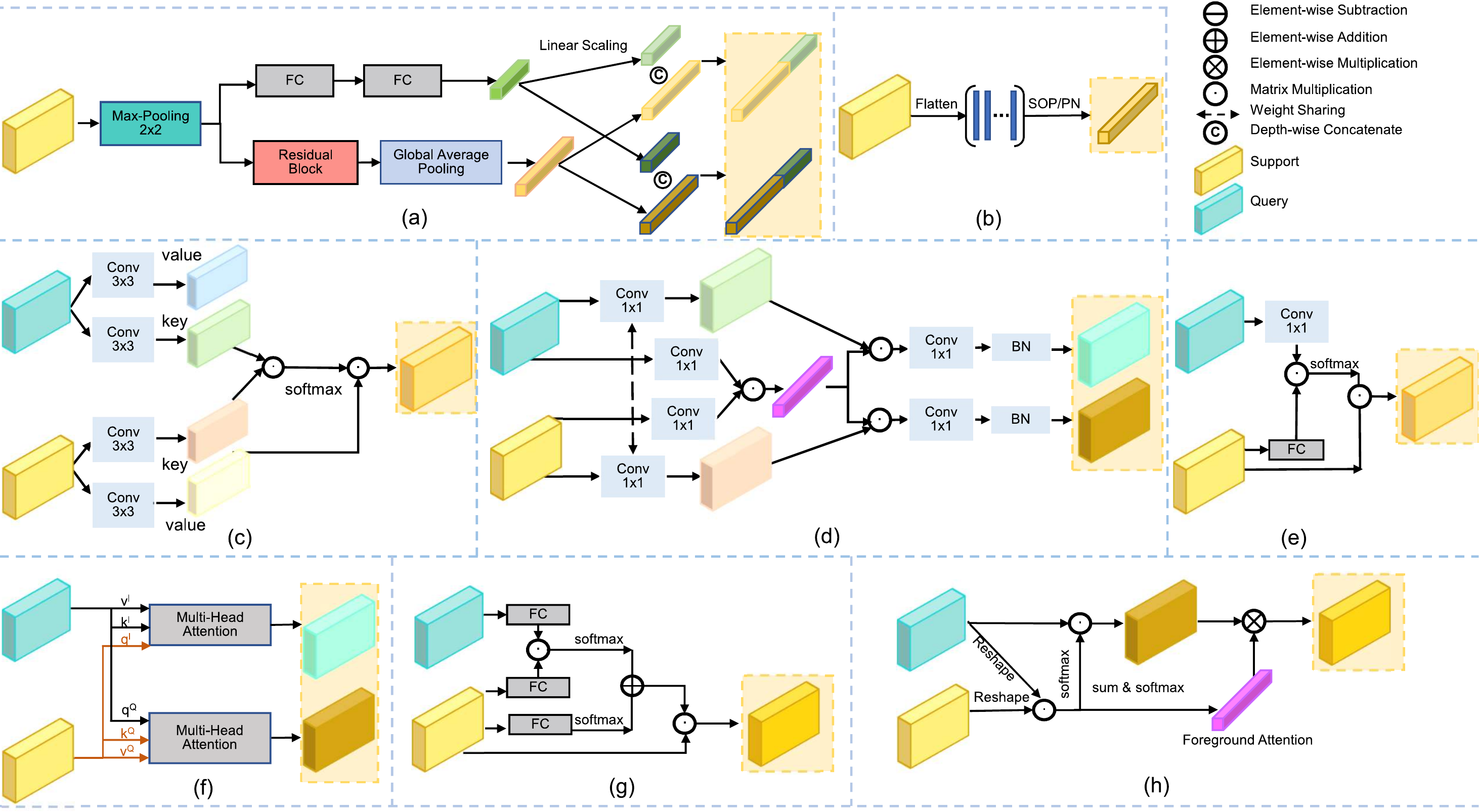}
\caption{The pipeline of support-only and support-query guidance. Especially, only (a) and (b) are support-only guidance while the other are support-query guidance. (a) Adaptive Fully-Dual Network (AFDN) \cite{liu2020afd}; (b) Hyper Attention RPN (HA-RPN) \cite{zhang2020few}; (c) Dense Relation Distillation (DRD) \cite{hu2021dense}; (d) Non-Local Co-Attention (NLCA) \cite{hsieh2019one}; (e) Augmentation with Conditioned Prototypes (ACP) \cite{wu2021universal}; (f) Multi-head Co-Attention (MHCA) \cite{chen2021adaptive}; (g) Cross-Image Spatial Attention Block (CISA) \cite{chen2021should}; (h) Spatial Alignment \& Foreground Attention Module (SAFA) \cite{han2021meta}.}
\label{fig_guidance}
\end{figure*}

\item{\textbf{Support-Only vs Support-Query Guidance.} To better integrate category-specific information into query features, there were two main ways to refine category prototypes, i.e., support-only and support-query guidance. In support-only guidance, it only takes category prototypes to refine features. Clearly, simple global max/average pooling was a kind of support-only guidance to capture global semantics which was helpful for the classification task \cite{kang2019few,yan2019meta,deng2020few}. However, it also lost spatial information (e.g., local structures) which was critical for RPN to make high-quality proposals. An effective way was to view a set of category prototypes as kernels to slide over query features for feature fusion. Liu et al. \cite{liu2020afd} designed a self-attention based fully-dual network that consisted of two parallel branches to simultaneously capture global and local semantics to generate task-specific attention vectors (Fig. \ref{fig_guidance}(a)). Zhang et al. \cite{zhang2020few} designed a list of PN (e.g., SigmE) to filter irrelevant factors in category prototypes (Fig. \ref{fig_guidance}(b)). 

In fact, objects in the support set usually have large differences with that in the query set, e.g., various postures and viewpoints. Meanwhile, support-only guidance does not take the misalignment problem between query and support features into consideration. Thus, the scoring function could easily confuse whether the RoI features are consistent with the prototypes of the associated category. To tackle the misalignment problem, support-query guidance uses the affinity matrix between query and support features to align features. As illustrated in Fig. \ref{fig_guidance}(c, e, g, h), these methods followed such a framework that first employed two parallel branches to get key and value maps for both query and support images, then took key maps to calculate an affinity matrix $A$ between query and support images, and finally applied the affinity matrix to align value maps of support to enhance value maps of query. Especially, it could be viewed as a search process that we search all spatial location of key maps of support to aggregate support evidence for feature fusion (like the mutual information). To refine the affinity matrix, Chen et al. \cite{chen2021should} presented a new branch which applied a fully-connected layer with spatial softmax to generate a pseudo foreground mask, which was used to reweight the affinity matrix, to filter background information. Similarly, Han et al. \cite{han2021meta} leveraged the amount of category evidence (i.e., sum up the affinity matrix by row) to determine each spatial location in query features whether or not a foreground region. Meanwhile, in Fig. \ref{fig_guidance}(d, f), several works explored a new paradigm which the affinity matrix was applied not only to align value maps of support to enhance value maps of query but also to align value maps of query to enhance value maps of support. In fact, such a scheme utilized a kind of the mutual relationship that both query and support objects would fetch enough information from each other if they were the same category while they would fetch very little from each other if They were of a different class. Thus, it could amplify the gap between the consistent pair and the other inconsistent pairs. Hsieh et al \cite{hsieh2019one} shared the affinity matrix between two parallel branches of query and support while Chen et al. \cite{chen2021adaptive} took separate affinity matrixes for each branch. Zhang et al. \cite{zhang2021accurate} designed a QSW module to build a category prototype from multiple support objects of a given category by their relevance to a query RoI.}

\item{\textbf{Aggregator.} As for feature fusion, there existed many excellent feature aggregators, e.g., channel-wise multiplication (MUL), element-wise subtraction (SUB) and channel-wise concatenation (CAT). MUL was commonly known as feature reweighting to learn feature co-occurrence. SUB was a kind of distance metric (i.e., L1 distance) to measure L1 similarity. CAT was a special aggregator that could stack features along the channel axis for subsequent networks to automatically explore a good way for feature fusion. Most existing methods only adopted MUL as their feature aggregator. Nevertheless, Han et al. \cite{han2021meta} and Liu et al. \cite{liu2020afd} combined MUL, SUB and CAT for better feature fusion (Fig. \ref{fig_feature_fusion}). In addition to traditional aggregators, another way was to transform category prototypes into a set of parameters, which were employed to initialize all weights of operators (e.g., a convolution operator), to enhance query features \cite{zhang2021accurate}.}

\begin{figure}[!t]
\centering
\includegraphics[width=0.5\textwidth]{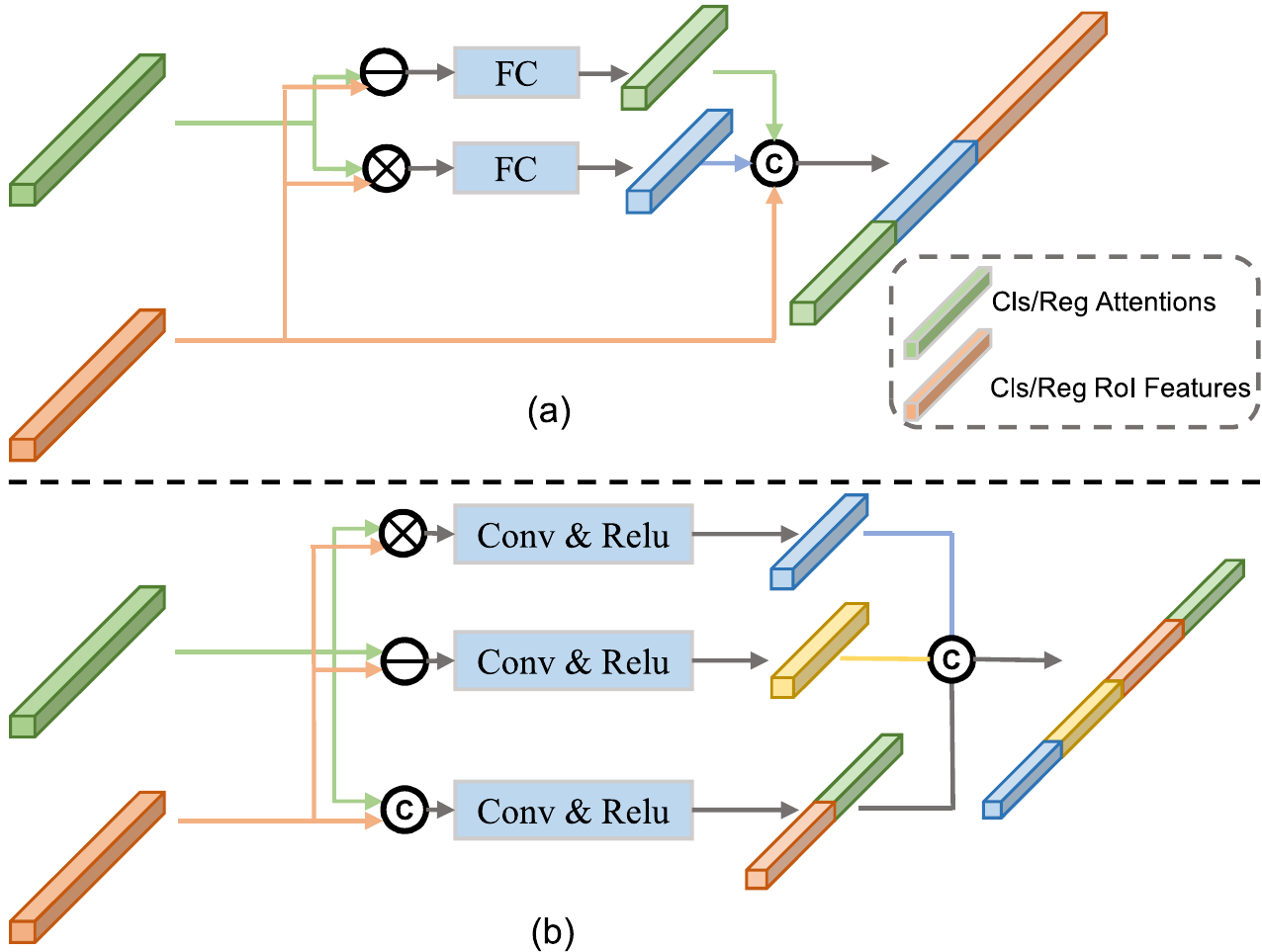}
\caption{The pipeline of feature fusion. (a) Dual Feature Aggregation; (b) Feature Fusion Network.}
\label{fig_feature_fusion}
\end{figure}

\begin{figure}[!t]
\centering
\includegraphics[width=0.5\textwidth]{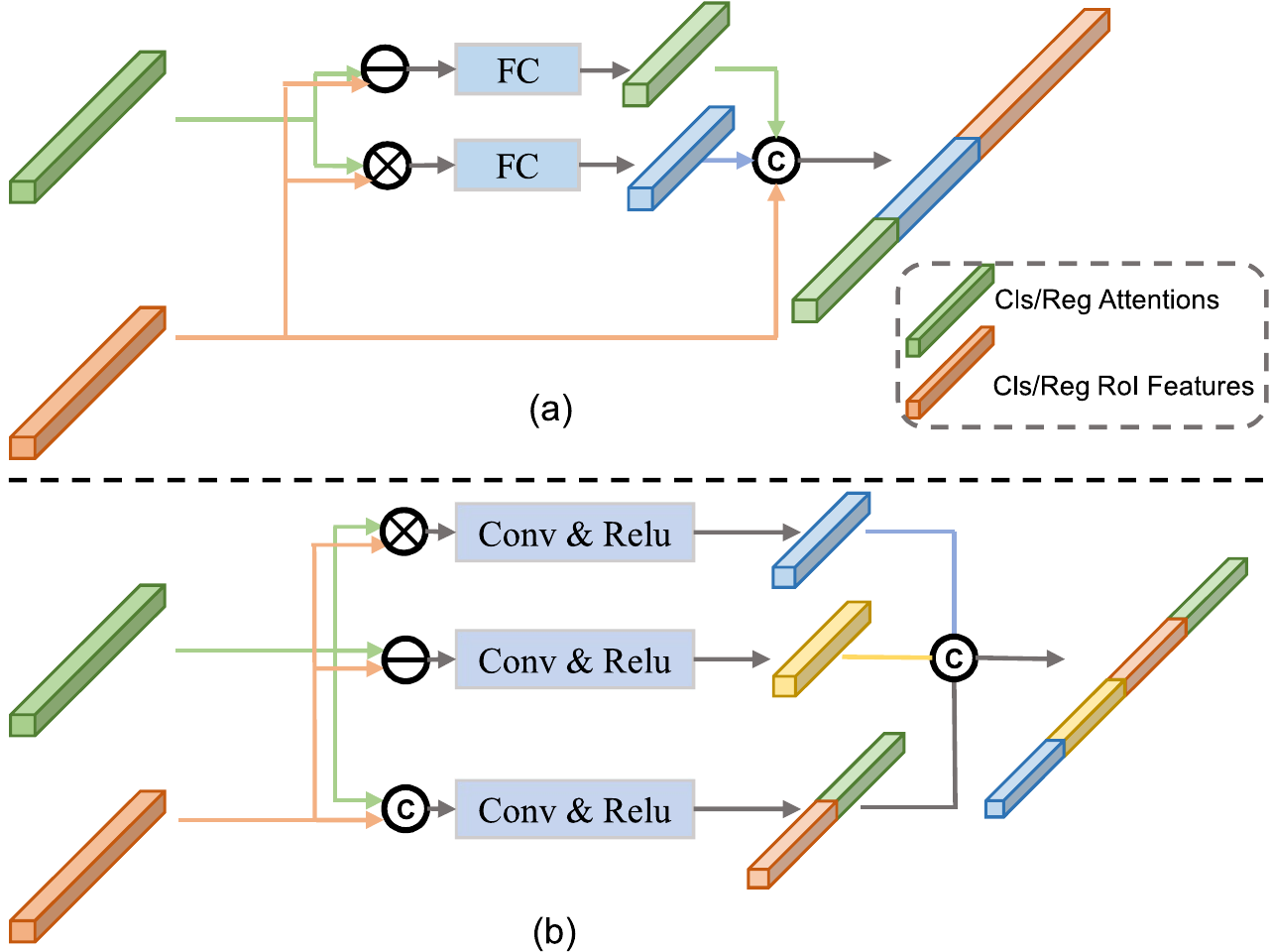}
\caption{The pipeline of variants of RNs \cite{sung2018learning}. (a) A Two-Layer MLP Relation Network; (b) Multi-Relation Network.}
\label{fig_RNs}
\end{figure}

\item{\textbf{Scoring Function.} A scoring function is another essential element for a metric-based method, including cosine similarity \cite{bansal2018zero}, Pearson similarity \cite{li2020mm} and variants of relation networks (RNs) \cite{liu2020afd,fan2020few,chen2021adaptive}. Especially, cosine and Pearson similarity were fixed metrics which may not fit a given task. To give a flexible metric for FSOD, variants of RNs learned a soft metric to recognize various categories. Most existing methods followed the one-vs-rest strategy to design a binary classifier (named the mete classifier) which should effectively distinguish whether RoI features were consistent with a set of prototypes of a given category or not (Fig. \ref{fig_RNs}(a)). A multi-relation network was designed to capture global-, local- and patch-relation between RoI features and class prototypes for overall matching (Fig. \ref{fig_RNs}(b)) \cite{yan2019meta,zhang2020few}. Zhang et al. \cite{zhang2021meta} regarded the scoring function as the transformer decoder to classify each query. 

Unlike the task of image recognition, an object detection method should distinguish negative proposals from all proposals, especially while negative proposals could even have a certain overlap (i.e., similarity) with associated class prototypes which may confuse the scoring function. To some extent, it was hard to define a background category to represent various negative samples with large intra-class variations. To tackle this problem, YOLO variants \cite{kang2019few,li2021beyond} took the confidence score to indicate the associated box whether or not a positive box. As aforementioned, R-CNN variants \cite{liu2020afd,fan2020few} employed a binary classifier to measure the similarity between RoI features with a set of prototypes for each class, where an RoI was assigned with a class tag that had the highest similarity score with this RoI. If the highest score was lower than a pre-defined threshold, it would be defined as a negative proposal. Yang et al. \cite{yang2020restoring} used negative proposals which had low IoU with ground-truth boxes to learn a list of negative prototypes for the background category. Bansal et al. \cite{bansal2018zero} explored two kinds of background definitions. First, a fixed vector was utilized as the prototype for the background class (statically assigned background). However, it could not really encode all background information, which could thus produce more false-positive samples. Second, in each iteration, we labeled background boxes with pseudo labels randomly sampled from a list of unseen categories to distinguish negatives from other positive proposals. Li et al. \cite{li2020mm} randomly sampled boxes, which had no overlap with ground-truth boxes, as exemplars of the background category to filter negative proposals of RPN.}

\item{\textbf{Loss Function.} A well-defined loss function was essential to learn a good model for a metric-based method. Many works inherited the loss functions from generic object detection methods, which were originally designed to guide the detectors to accurately detect all objects in an image. However, it did not take class prototypes into account, which played a key role in a metric-based method. Thus, several works introduced the customed loss function which could guide the embedding network to simultaneously minimize intra-class variations of prototypes and maximize inter-class distance of prototypes. Yan et al. \cite{yan2019meta} appended an MLP with softmax to the embedding network and employed cross-entropy (CE) loss to encourage the embedding network to explicitly encode category information into category prototypes. Fan et al. \cite{fan2020few} presented a two-way contrastive training strategy. It first constructed a training triplet $(q,s_c,s_n)$ where $q$ was a query image and $s_i$ was a support instance of the i-th category ($c\neq n$). Then, it combined $s_c$ and $q$ to generate category-specific features fed to RPN to generate a set of proposals $p$, where only proposals of the c-th category were regarded as positive proposals. Next, it proportionally sampled training pairs to construct a balanced training set, consisting of three types of pairs, i.e., $(p_c,s_c)$, $(p_b,s_c)$ and $(p_\ast, s_n)$ where $p_i$ was a proposal of i-th category ($b \neq c$). Finally, it enforced a binary cross-entropy (BCE) loss to learn a good representative. Margin-based ranking loss \cite{chen2021adaptive} was a kind of multi-task loss, consisting of two parts. One was a hinge loss variant for the foreground-background classification. The other was a max-margin contrastive loss to enforce all RoIs to satisfy max-margin category separation and semantic space clustering as possible. Bansal et al. \cite{bansal2018zero} defined a margin loss to enforce a constraint that the matching score of a proposal with its true category should be higher than that with other categories. Li et al. \cite{li2021beyond} assumed that novel classes had implicit relation with base classes and prototypes of novel classes could be embedded into margins between prototypes of base classes. However, large inter-class distance provided the safe decision boundary for the classification while large margin between prototypes of base class makes it hard for novel classes to find appropriate class prototypes. Thus, Li et al. \cite{li2021beyond} proposed a max-margin loss which was formulated as a sum of intra-class distance of all classes over a sum of inter-category distance of all classes to adaptively adjust class-margin. Zhang et al. \cite{zhang2021accurate} improved traditional contrastive loss from two aspects. First, learnable margins were designed to adaptively adjust inter-class distance. Especially, these margins were initialized by inter-class semantic similarity. Second, instead of hard sampling in the basic contrastive loss, focal loss was imposed to adaptively adjust contributions of various kinds of samples to the gradient. Li et al. \cite{li2021transformation} also proposed a transformation invariant principle to learn a robust embedding network which should produce consistent category prototypes between an image and an associated transformed image, except for CE loss.}

\item{\textbf{Fusion Node.} As described above, class prototypes encoded class-specific information, e.g., the shape of objects, which were helpful for the detectors to accurately localize and classify objects of the associated category. Thus, it was important to determine where to place fusion nodes to aggregate query features and class prototypes in one-/two-stage detectors. Tab. \ref{tab_fusion_node} presented common fusion nodes and affected modules of one- and two-stage detectors. 

For a two-stage detector, many works have explored six fusion nodes, in Fig. \ref{fig_fusion_nodes}(a). Several works revised the multi-class classifier as the meta-classifier which combined RoI features and category prototypes of a given category as inputs and output a matching score, i.e., only the 5th fusion node \cite{bansal2018zero, li2020mm,karlinsky2019repmet,zhu2021semantic}. However, it ignored the implicit shape and position information in category prototypes, which had a positive effect on the bounding box regressor. Thus, the 4th or 6th fusion node was widely added to provide category-sensitive information for the R-CNN head \cite{liu2020afd, li2021beyond, li2021transformation, yan2019meta}. Due to differences between the classification and localization, Liu et al. \cite{liu2020afd} explicitly decomposed the 4th fusion node into the 5th and 6th feature node which provided task-specific information respectively for two sub-tasks. In addition, to get more high-quality proposals from RPN, many works inserted the 1st or 2nd fusion node to provide class-sensitive information for RPN (meta-RPN) \cite{fan2020few,chen2021should,li2020one}. The 1st fusion node could affect the whole Faster R-CNN head while the 2nd fusion node only served RPN \cite{zhang2021meta,hu2021dense}. For one-stage detectors, almost all methods chose the 1st fusion node to make better results under the help of category-sensitive information (Fig. \ref{fig_fusion_nodes}(b)) \cite{deng2020few, li2021beyond,kang2019few}. More details about combinations of fusion nodes were shown in Tab. \ref{metric_learning_all_methods_review}.}

\begin{table}[!t]
  \centering
  \caption{Modules Affected by Fusion Node. Note that a-i is the i-th fusion node of two-stage detectors while b-i is the i-th fusion node of one-stage detectors, as in Fig. \ref{fig_fusion_nodes}.}
    \centering
\begin{tabular}{|c|c|}
\hline
Fusion Nodes & Affected Modules            \\ \hline
a-1          & RPN \& Cls \& Loc           \\ \hline
a-2          & RPN                         \\ \hline
a-3          & \multirow{2}{*}{Cls \& Loc} \\ \cline{1-1}
a-4          &                             \\ \hline
a-5          & Cls                         \\ \hline
a-6          & Loc                         \\ \hline
b-1          & Cls \& Loc                  \\ \hline
b-2          & Cls                         \\ \hline
b-3          & Loc                         \\ \hline
\end{tabular}
\label{tab_fusion_node}
\end{table}

\begin{figure}[!t]
\centering
\includegraphics[width=0.5\textwidth]{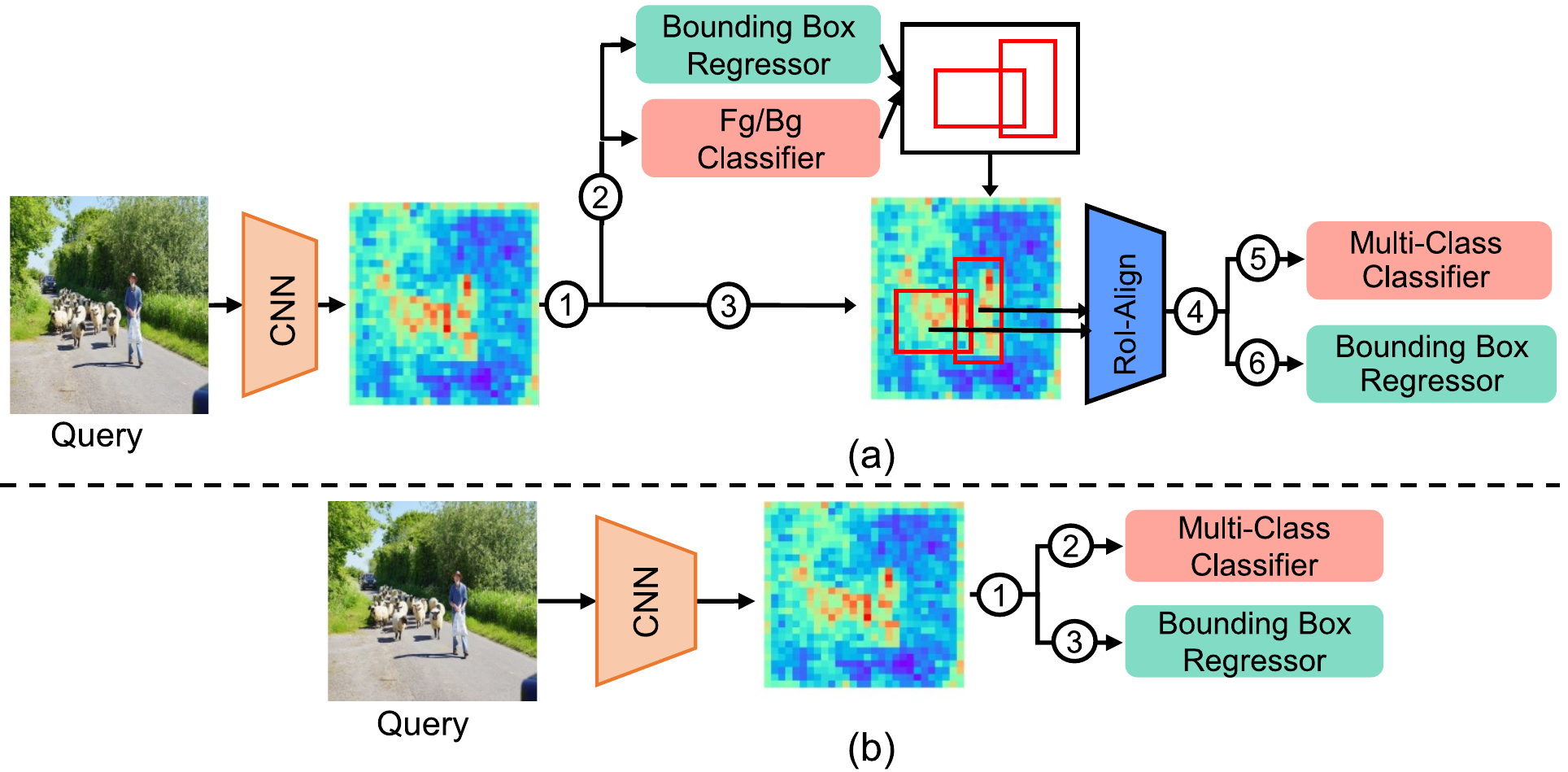}
\caption{Fusion Nodes of two parallel branches of support and query in (a) two- stage detectors/(b) one-stage detectors. Note that \textcircled{i} represents the i-th fusion node which is actually a feature aggregator to aggregate query features and category prototypes.}
\label{fig_fusion_nodes}
\end{figure}

\item{\textbf{Training/Testing Process.} The overall training/testing process is illustrated in Algorithm \ref{alg:Framwork}, i.e., meta-training/finetuning/testing. Almost all metric-based methods are usually pretrained in existing benchmarks \cite{chen2015microsoft,deng2009imagenet} to learn basic notions (e.g., low-level visual features) which could be applied into our $N$-way $K$-shot problem, instead of training from scratch. For a fair comparison, it is essential to remove all categories in ImageNet overlapped with unseen categories in target benchmarks. In meta-training stage, we first construct a set of tasks/episodes built upon $D_{base}$ to let $M^{base}$ learn how to leverage limited reliable data to match all instances in a query set. In other words, a base learner is actually a task-agnostic learner which could be instantiated with task-specific notions from a given support set to solve a novel task. Thus, high-quality support data is crucial to learn a robust $M^{base}$. Tab. \ref{strategy_for_stable_learning} presents several solutions to pair a relatively appropriate support for a query image when plenty of small/hard objects are in $D_{base}$. Especially, meta-finetuning is an optional step in a metric-based method. In most cases, meta-finetuning could further improve the learner’s performance in meta-testing stage. However, it is unrealistic to directly leverage $D_{base} \cup D_{novel}$ to finetune $M^{finetune}$ owing to the extreme class imbalance between $D_{base}$ and $D_{novel}$. In general, we need to construct a balanced subset which is randomly sampled from $D_{base}$ ($K_b$ exemplars per category). In most cases, $K_b$ is equal to $K$ ($N$-way $K$-shot). However, considering a large number of training exemplars for base classes, $K_b$ can be increased properly to get more robust prototypes for base classes to speed up the training process, e.g., $K_b=3K$ \cite{yan2019meta}. In meta-testing, we could preprocess $D_{novel}$ to get class-sensitive vectors for each novel class as task-specific parameters for $M^{finetune}$ instead of temporarily extracting for all tasks which is time-consuming. Due to low-shot scenarios, base/novel-category splitting settings have a great effect on final results. Thus, two main ways are proposed to evaluate performance of $M^{finetune}$ more accurately, in Tab. \ref{evaluation_settings}.}

\begin{table}[!t]
  \centering
  \caption{Solutions to Construct a More Suitable Episode from A Difficult Benchmark.}
    \begin{tabularx}{\linewidth}{|c|X|}
    \hline
    Solution & \multicolumn{1}{c|}{Explanation} \\
    \hline
    Hsieh et al. \cite{hsieh2019one} & Take a pretrained Mask R-CNN to evaluate the difficulty of a ground-truth box and only ground-truth boxes with high-quality proposals would be cropped out as support patches. \\
    \hline
    Fan et al. \cite{fan2020few} & Remove all images with a box whose size is less than $32 \times 32$ in the training stage. \\
    \hline
    Li et al.\cite{li2020one} & Randomly sample $m$ support patches and only choose a support patch which has the most similarity with the query images. \\
    \hline
    Liu et al. \cite{liu2020afd} & Sample more images to build a larger support set for each task in $\mathcal{T}^{train}$ ($K=200$). \\
    \hline
    \end{tabularx}
      \label{strategy_for_stable_learning}
\end{table}

\begin{table}[!t]
  \centering
  \caption{Details of Two Main Evaluation Settings.}
    \begin{tabularx}{\linewidth}{|c|X|}
    \hline
    Method & \multicolumn{1}{c|}{Detail} \\
    \hline
    Karlinsky et al. \cite{karlinsky2019repmet} & Randomly sample 500 episodes ($n$-way $k$-shot tasks) where each episode consists of a support set with $k$ instance per category ($|S|=nk$) and a query set $Q$ with 10 images per category ($|Q|=10n$). Thus, we have at least 10 instances of each category in meta-testing stage. ($n$ is usually assigned by 5.) \\
    \hline
    Fan et al. \cite{fan2020few} & (1) Randomly sample several base and novel class splitting settings;\newline{} (2) Construct a list of tasks $\mathcal{T}^{test}$ built upon $(D_{novel},D_{novel}^\prime)$ for each splitting setting to evaluate performance;\newline{} (3) Report average performance for each splitting setting. \\
    \hline 
  \end{tabularx}%
  \label{evaluation_settings}
\end{table}%

\begin{table}[!t]
  \centering
  \caption{Summarization of details of representative methods for two other settings of support branches.}
    \begin{tabularx}{\linewidth}{|c|c|X|}
    \hline
    Group & Method &\multicolumn{1}{c|}{Explanation} \\
    \hline
    \rotatebox[origin=c]{90}{Learnable Visual Prototype (LVP)} & Yang et al. \cite{yang2020restoring} & Meta-Training: (1) Assign $K$ learnable positive \& negative prototypes per base category; (2) Randomly initialize all prototypes; (3) Leverage $D_{base}$  to create $n$-way $k$-shot episodes (sample $n$ support categories instead of using all categories for computational efficiency) to learn all trainable parameters.\newline{}Meta-Testing: (1) Initialize task-specific parameters: (a) Novel-Category Prototypes: Use weight imprinting to initialize novel-category prototypes; (b) Background-Category Prototypes: Apply spectral clustering on embedding vectors of $M$ hard negative prototypes to generate $K$ negative prototypes respectively for each novel category.; (2) Evaluate NP-RepMet on $D_{novel}^\prime$. \\
    \hline
    \rotatebox[origin=c]{90}{Semantic Prototype (SP)} & Bansal et al. \cite{bansal2018zero} & Meta-Training: (1) Prepare Training Data: (a) Exploit EB \cite{zitnick2014edge} to generate raw proposals; (b) Sample both positive and negative proposals as training data; (c) Assign base class labels to positive proposals; (d) Randomly sample a list of negative classes that have no overlap with these base classes and assign these negative class to negative proposals; (2) Take a word vector as prototypes for each base and negative class; (3) Exploit LAB to learn a meta detector.\newline{}Meta-Testing: (1) Use a word vector as the category prototype for each novel category; (2) Evaluate ZSD on $D_{novel}^\prime$. \\
    \hline
    \end{tabularx}%
    \label{other_setting_for_support}
\end{table}%

\begin{algorithm}[!t]
  \scriptsize
   \caption{Meta-Training/Finetuning/Testing of A Metric-based Method}
  \label{alg:Framwork}
  \begin{algorithmic}[1]
    \Require
     \Statex A pre-defined model, $M=\{M^{train}, M^{finetune}\}$;
      \Statex A large-scale benchmark for base classes, $D_{base}$;
      \Statex A small-scale training benchmark for novel classes (N samples per class), $D_{novel}$;
      \Statex A testing benchmark for novel classes, $D^\prime_{novel}$; // $D^\prime_{novel}\cap D_{novel} = \emptyset $
\end{algorithmic}
    \par\vspace{-0.5\baselineskip}\noindent\hrulefill \\
\textcolor{red}{Meta-Training}
\begin{algorithmic}[1]
    \State Initialize all parameters of $M^{train}$;  
    \For{$i \leftarrow 0, 1, ..., e_0-1$} // $e_0$: the number of epochs in meta-training;
    \State Construct a set of tasks/episodes $\mathcal{T}^{train}$ built upon $D_{base}$;
    \For{$T$ in $\mathcal{T}^{train}$}
    \State $l^{train} \leftarrow Loss^{train}(M^{train}, T)$
    \State Update all trainable parameters $\theta^{train}$ of $M^{train}$ by backprop;
    \EndFor
    \EndFor
 \end{algorithmic}
\textcolor{red}{Meta-Finetuning}
\begin{algorithmic}[1]
    \State Initialize all parameters of $M^{finetune}$; // shared parameters copied from $M^{train}$;
    \For{$i \leftarrow 0, 1, ..., e_1-1$} // $e_1$: the number of epochs in meta-finetuning;
    \State Construct a set of tasks $\mathcal{T}^{finetune}$ built upon $D_{base} \cup D_{novel}$;
    \For{$T$ in $\mathcal{T}^{finetune}$}
    \State $l^{finetune} \leftarrow Loss^{finetune}(M^{finetune}, T)$
    \State Update all trainable parameters $\theta^{finetune}$ of $M^{finetune}$ by backprop;
    \EndFor
    \EndFor
 \end{algorithmic}

\textcolor{red}{Meta-Testing}
   \begin{algorithmic}[1]
    \State Construct a set of tasks $\mathcal{T}^{test}$ built upon $D_{novel} \cup D^\prime_{novel}$; //  $(S_i, Q_i) = T_i \in \mathcal{T} \& S_i \subset D_{novel} \& Q_i \subset D^{\prime}_{novel} \& Q_1 \cup Q_2 \cup ... = D^{test}$
   \State $ans \leftarrow \{\}$
    \For{$T$ in $\mathcal{T}^{test}$}
    \State $bboxes=M^{finetune}(T)$
    \State $ans \leftarrow ans \cup \{(Q, bboxes)\}$
    \EndFor
   \State Take $ans$ to evaluate performance of $M^{finetune}$ by a metric, e.g., mAP50.
  \end{algorithmic}
\end{algorithm}

\item{\textbf{Other Settings of Support Branch.} Earlier, we mainly discussed a kind of method that employed a list of support images fed to a customed backbone for category prototypes named Generative Visual Prototype (GVP). Currently, two other settings of support branches have been proposed to get rid of sampling a set of support images per category for all episodes, i.e., LVP and SP, as illustrated in Tab. \ref{other_setting_for_support}. To mitigate noise in a small support set, LVP aims to exploit a list of learnable kernels to automatically acquire class prototypes for each class, without dependency on a specific object. Several works \cite{bansal2018zero} explored to use of semantic prototypes (SPs) learned from a large-scale corpus to provide reliable task-specific parameters for these detectors. }
\end{itemize}

\emph{2. Optimization-based Methods.} Optimization-based methods assume that generic notions (i.e., meta parameters and a meta optimizer) could be learned from the base datasets to provide a suitable gradient guidance or a uniformly optimal initial weight for quick adaption to a new task. However, unlike FSC, it is difficult for a meta-optimizer or meta-parameters to adapt to a wider parameter space or to balance two sub-tasks of object detection. 
Therefore, optimization-based learning has very few applications in LS-FSOD.

\begin{itemize}
\item{\textbf{Meta RetinaNet}, proposed by Li et al. \cite{li2020meta}, was an optimization-based detector which took a MAML variant \cite{finn2017model} to learn meta-parameters for a new task. To reduce too large parameter space, kernels $k$ of convolution layers in RetinaNet was reformulated as: $k^\prime = k \odot w$, where $w$ were learnable coefficient vectors initialized by ones and $k$ were constant kernels initialized by associated kernels pretrained on $D_{base}$. In the meta-training phase, balanced loss (BL) was designed to replace simple summation in original MAML to adaptively down-weight easy tasks and focus on hard tasks to update meta parameters (i.e., coefficient vectors and parameters of the last classification layer). }
\end{itemize}

\emph{3. Model-based Methods.} For model-based methods, the key is to design a model and associated learning strategies which quickly adapts for a new episode. Most model-based methods mainly relied on RNNs \cite{zaremba2014recurrent} with memory to utilize training samples of an episode to predict task-specific parameters for such an episode \cite{cai2018memory,santoro2016meta}. Parameters could be further divided into fast/slow (task-specific/agnostic) parameters which were combined to make more stable predictions. However, like optimization-based learning, model-based learning was barely noticed in FSOD.

\begin{itemize}
\item{\textbf{MetaDet.} Inspired by a category-agnostic transformation, Wang et al. \cite{wang2019meta} first explored a paradigm for a meta generator $G$: $w^\ast=G(w)$, where $w$ and $w^\ast$ were task-specific parameters learnt respectively from an episode and a large-scale dataset. In MetaDet, only R-CNN head was defined as task-specific modules. Especially, its training consisted of three stages.

1. Pre-Training Phase: Take $D_{base}$ to learn a large-sample detector $D(I;\theta^\ast)$ in a standard way, where $\theta^\ast$ is a set of parameters of the detector.

2. Meta-Training Phase: First, learn a $n$-way $k$-shot detector $D(I;\theta^\ast \cup w_{det}^{c_{base}} \backslash w_{det}^{c_{base},\ast})$ where $w_{det}^{c_{base},\ast}$ and $w_{det}^{c_{base}}$ represent category-specific parameters learned in $D_{base}$ or a $n$-way $k$-shot episode in the top layer of Faster R-CNN. Then, add consistency loss on the pair $(w_{det}^{c_{base},\ast},w_{det}^{c_{base}})$ of each episode to learn a meta generator G.

3. Meta-Testing Phase: Fist, learn a $n$-way $k$-shot detector $D(I;\theta^\ast \cup  w_{det}^{c_{novel}} \backslash w_{det}^{c_{base},\ast})$ where $w_{det}^{c_{novel}}$ denote category-specific parameters learned in $D_{novel}$. Then, feed $w_{det}^{c_{novel}}$ to G for a more robust version $w_{det}^{c_{novel},\ast}$. Finally, fine-tune $D(I;\theta^\ast \cup w_{det}^{c_{novel},\ast} \backslash w_{det}^{c_{base},\ast})$.

However, MetaDet mainly relied on a principle that base categories shared similar distribution with novel categories, which greatly limited its scope of applications.}
\end{itemize}

\textbf{Transfer-Learning Methods.}
Compared with meta-learning algorithms, transfer-learning methods usually have two phases of training, i.e., the pre-training and fine-tuning stage. In the pre-training stage, a large-scale dataset $D_{base}$ is employed to train a base detector for general notions under the official setting. There are two steps during the fine-tuning stage: (1) Initialization. The novel learner inherits general notations/parameters from the base learner while the rest parameters are initialized randomly or by a weight imprinting technique. (2) Fine-tuning. A small-scale dataset $D_{base}^\prime\cup D_{novel}$ is constructed to fine-tune the novel learner where $D_{base}^\prime$ is a balanced subset of $D_{base}$ ($K$ shots per base class). To sample a more suitable subset $D_{base}^\prime$, Li et al. \cite{li2021class} proposed a clustering-based exemplar selection algorithm which first calculated intra-category mean features/prototypes for each image in $D_{base}$, then applied the k-means algorithm to generate $K$ clusters for each base category, and finally obtained $K$ centroids (exemplars) per base category. Like aforesaid metric-based learning, we will make a detailed review on the fine-tuned module, regularization, classifier and loss function for transfer-learning methods. If you want to know specific contributions, we also present them in Tab. \ref{details_transfer_learning}. 

\begin{itemize}
\item{\textbf{Fine-tuned Module.} In the early stage of transfer-learning approaches \cite{wang2020frustratingly}, the backbone and RPN were usually taken as task-agnostic modules and it was ineffective to utilize a small-scale dataset $D_{base}^\prime \cup D_{novel}$ to fine-tune all parameters of a base detector. Thus, wang et al. \cite{wang2020frustratingly} explored a FIX\_ALL mode which only adapted the final layer of both classification and localization in the fine-tuning stage and achieved promising results. However, as aforesaid in Section \ref{subsec_challenges}, RPN showed relatively poor performance for novel classes. Sun et al. \cite{sun2021fsce} demonstrated the number of positive proposals for novel categories was about a quarter of that for base categories. Hence, there were not enough positive proposals to fine-tune RPN in the fine-tuning stage. Meanwhile, it implicitly introduced the foreground-background imbalance problem, due to low-quality proposals for novel categories. Sun et al. \cite{sun2021fsce} proposed a two-step procedure to re-balance sampling ratio of the positive and negative proposals: (1) reuse positive proposals with low confidence suppressed by NMS; (2) discard negative proposals by half. Moreover, backbones were proved to have weaker response for novel categories than that for base categories. Thus, many works explored an extreme mode (named FIT\_ALL) to obtain a de-biased backbone/FPN for better features \cite{wu2021universal,wu2020multi,wang2019few}. However, low density sampling easily leaded to overfitting under the mode FIT\_ALL.}

\begin{figure}[!t]
\centering
\includegraphics[width=0.4\textwidth, height=0.4\textwidth]{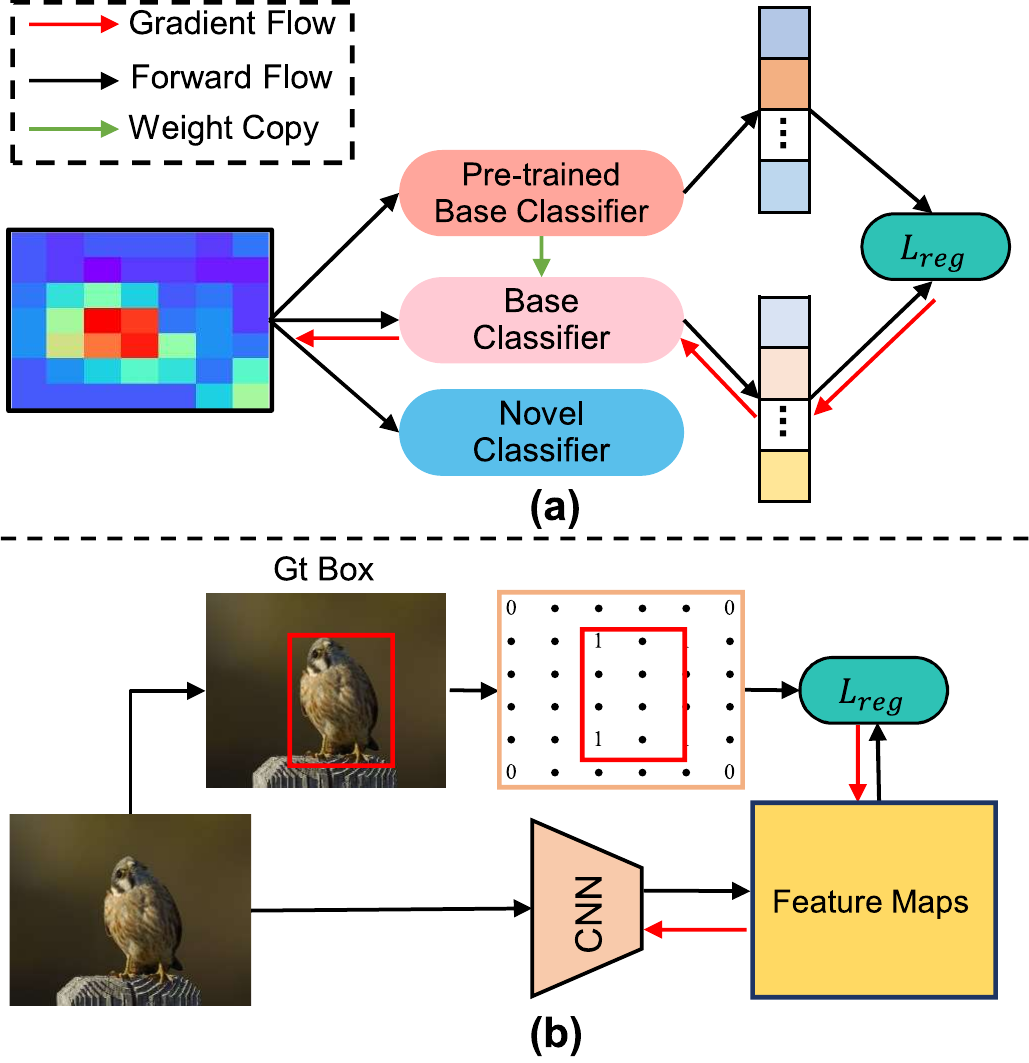}
\caption{The pipeline of two classic regularization methods. (a) Transfer-Knowledge (TK) Regularization; (b) Background-Depression (BD) Regularization.}
\label{fig_reg}
\end{figure}

\begin{table}[!t]
  \centering
  \caption{The sub-modules of FAFRCNN.}
    \begin{tabularx}{\linewidth}{|c|X|}
    \hline
    Sub-Module & \multicolumn{1}{c|}{Detail} \\
    \hline
    \makecell{Split \\ Pooling} & (1) Apply shared random shifts on a set of pre-defined anchors to crop multi-scale features patches for both source and target domain.\newline{}(2) Take feature patches to generate two kinds of pairs at each scale (i.e., source-source and source-target pairs).\newline{}(3) Adopt generative adversarial learning to alternatively optimize the generator (backbone) and the discriminator that classifies feature patches whether or not belong to target domain. \\
    \hline
    \makecell{Instance-level \\ Adaptation} & (1) Sample foreground RoIs with high IoU threshold (i.e., 0.7) in source and target domain. \newline{}(2) Take RoI features to generate two kinds of pairs (i.e., source-source and source-target pairs).\newline{}(3) Adopt generative adversarial learning to alternatively optimize the generator (RoI feature encoding network) and the discriminator that classifies RoI features. \\
    \hline
    \end{tabularx}%
    \label{details_FAFRCNN}
\end{table}%

\item{\textbf{Regularization.} To alleviate overfitting, several studies added extra constraints on the parameter space for a robust model. Earlier, the base classifier was usually dropped in the fine-tuning stage \cite{sun2021fsce,wang2020frustratingly,wu2020multi}. Especially, it could be viewed as a special encoder that took RoI features $R$ as input and output a similarity score for each pair in a set $\{(R,C_{base}^i)|i=1,2,\ldots,|C_{base}|\}$. In other words, it could be utilized to indicate shared features among novel and base classes to regularize the models. Thus, as shown in Fig. \ref{fig_reg}(a), KL loss was applied to enforce the model to reuse shared features for a novel task \cite{li2021class,wu2021universal,chen2018lstd}. Moreover, Wu et al. \cite{wu2021universal} leveraged a consistent constraint (KL loss) among probability distributions generated by a shared classifier and reserved a set of class-agnostic prototypes learnt from a large-scale dataset $D_{base}$ for a novel task as well for full use of generic notions from base classes.

Complex background may degenerate performance in a few-shot scenario. As shown in Fig. \ref{fig_reg}(b), Chen et al. \cite{chen2018lstd} utilized ground-truth boxes in a given image for making a rough estimation of the background mask to explicitly depress background features. In addition, several works leveraged a pre-trained saliency model to generate saliency maps to reweight features for background depression \cite{chen2020leveraging}.

FPN \cite{lin2017feature} was widely utilized to alleviate large scale variations. FPN employed several feature maps at different scales for region proposals. However, due to low-density sampling, it was hard to provide enough positive exemplars at each scale to train FPN \cite{lin2017feature} while more negative proposals were generated by FPN. Thus, Wu et al. \cite{wu2020multi} proposed object pyramids for more exemplars to refine FPN to distinguish positive-negative exemplars at each scale. Instead of directly extracting objects, a small random shift was applied on square boxes for some disturbance to learn a more robust FPN \cite{lin2017feature}. Then, we resized objects to six scales (i.e., $\{{32}^2,\ {64}^2,{128}^2,{256}^2,{512}^2,{800}^2\}$). Next, objects were fed into shared Faster R-CNN \cite{ren2015faster}. Finally, for objects at a special scale, only feature maps at an associated scale were leveraged to calculate RPN and RoI classification loss.

As described above, many works supposed that the novel classes $C_{novel}$ shared generic notions with the base classes $C_{base}$ (i.e., low domain shift). Nevertheless, it was hard to collect a proper dataset $D_{base}$ in some special cases. Hence, Wang et al. \cite{wang2019few} proposed an adversarial learning to mitigate domain shifts (Tab. \ref{details_FAFRCNN}).}

\item{\textbf{Classifier.} For most transfer-learning methods, error rates in classification are much higher than that in localization. Thus, several works replaced the softmax classifier with the cosine classifier to get rid of irrelevant factors, e.g., various feature norms. Moreover, Yang et al. \cite{yang2020context} designed Context-Transformer to provide more contextual information for the novel classifier. It first leveraged spatial pooling to implicitly extend receptive/contextual fields of original features $F_k$ at each scale $k\in\{1,2,\ldots,K\}$ for context features $Q_k$. Then, an affinity matrix $A_k$ was calculated among features $F_k$ \& a set of context features $Q=\{Q_1,Q_2,\ldots,Q_K\}$ to adaptively match effective supporting context at various scales for feature fusion.}

\item{\textbf{Loss Function.} Multi-class cross entropy (MCE) loss was generally used for classification. However, MCE aimed at increasing inter-class distinction which pay little attention to lower intra-class variations. Thus, Chen et al.\cite{chen2020leveraging} utilize a variant of cosine loss to form compact category cluster. Sun et al. \cite{sun2021fsce} introduced a contrastive learning strategy which leveraged a variant of cross entropy to introduce competition between homogeneous and heterogeneous pairs for high inter-category distinction and low inter-category distance.}
\end{itemize}

\emph{3. Data Augmentation Methods.}
Data augmentation aims at increasing intra-class variations to enforce the model to utilize more robust features. Li et al. \cite{li2021transformation} proved, in few-shot scenarios, most naive data augmentation methods could introduce large intra-class variations which had negative effect on its performance. Therefore, Li et al. \cite{li2021transformation} proposed a TIP module to construct contrasting pairs to make the encoder invariant to various intra-class variations by KL loss. Likewise, Zhang et al. \cite{zhang2021hallucination} designed a hallucination network to transfer the intra-category variations from base categories to novel categories. Simple horizontal flipping was proved effective for the performance improvement due to moderate variations \cite{li2020one}.

\subsection{Imbalanced Limited-Supervised Problem}
\subsubsection{Definition}
In a low-shot scenario, it is likely to collect a imbalanced dataset due to the long-tail distributions of real world data. Objects have great discrepancies in the occurrence frequencies (i.e., $\min{\{|D_{novel}^i|,\ i\in C_{novel}\}} \ll \max{\{|D_{novel}^i|, i\in C_{novel}\}}=k$). Moreover, it is also sub-optimal to adopt an under- or over-sampling strategy to build a $N$-way $K$-shot benchmark. However, such a problem has not attracted enough attention.

\subsubsection{Solutions}
So far, there were a few works \cite{zhang2020class} to present the detailed description on this imbalanced problem within our capacity. In addition, we also propose several candidate solutions for this problem, i.e., group-based methods and generative methods.

Inspired by focal loss, a solution could be implemented by a balanced loss which could adaptively re-balance gradient from all categories. Zhang et al. \cite{zhang2020class} proposed CI loss which employed the imbalance degree in a dataset to automatically select appropriate parameters for gradient re-balance. Zhang et al. exploited NUDT-AOSR15 to construct both the training and testing set and chose mAP as the evaluation metric.

Inspire by the long-tail distribution in the large-scale dataset (e.g., PASCAL VOC07/12), we propose two kinds of solutions that could be applied to tackle this imbalanced limited-supervised problem. First, GAN variants were introduced to produce high-quality exemplars (e.g., images) for re-balancing the foreground-foreground class  \cite{dwibedi2017cut,dvornik2018modeling,tripathi2019learning}. Second, a classification tree was built upon lexical or semantic relations for a coarse-to-fine strategy to alleviate class imbalance \cite{wu2020forest}, instead of directly classifying all classes. In addition, there are some tricks to tackle this problem, such as an NMS resampling algorithm \cite{wu2020forest} which dynamically adjusted NMS thresholds for each category to provide enough RoIs for the class with a few training exemplars according to the occurrence frequency of each category.

\section{Semi-Supervised Few-Shot Object Detection}
\label{sec_semi}
So far, most works on semi-supervised object detection (SSOD) collected about half (or 10 percent) of data with instance-level labels, which is fundamentally different from few-shot settings and still costs a lot. Especially, if there exist over 1000 images in a dataset for SSOD, SSOD will have far more labeled data than that annotated for SS-FSOD. Thus, compared with traditional SSOD, SS-FSOD can further reduce the annotation burden and exploit labeled data more effectively. There are two main ways for SS-FSOD, i.e., self-training and self-supervised based solutions.

\subsection{Problem Definition}
Let $D_{novel}$ represent a small dataset with a few labeled exemplars for a set of classes $C_{novel}$ and $D_{novel}^-$ represent a large-scale dataset sampled in the target domain without target supervision. For each sample $(I,Y)$ in $D_{novel}$, $I$ is an image ($I\in\mathbb{R}^{M\times N\times3}$) and $Y=\{(b_n,y_n)\}^N$ is a set of $N$ instances of $C_{novel}$ in $I$, where $b_n\in\mathbb{R}^4$ is a bounding box of the n-th instance and $y_n\in\{0,1\}^{|C_{novel}|}$ denotes an one-hot class encoding. In some cases, we also have an extra dataset $D_{base}$ without target-domain supervision to learn generic notions instead of training from scratch.

Let $D_{novel}^n$ be a set of instances of the n-th category in $D_{novel}$. Like in limited-supervised FSOD, we restrict the maximum of the number of instances per category in $D_{novel}$: $\max{\{|D_{novel}^i|,\ i\in C_{novel}\}}\le k$ ($k$ is usually no more than 30). In all, SS-FSOD tries to extract latent notions in $D_{novel}^-$ (e.g., pseudo labels) to refine/pre-train detectors for avoiding overfitting in $D_{novel}$.

\subsection{Dataset}
In SS-FSOD, it mainly uses existing benchmarks to form a semi-supervised few-shot dataset, including PASCAL VOC07/12, MSCOCO and ImageNet-LOC. We usually take mAP for PASCAL VOC07/12 and mAP/mAP75 for MSCOCO to evaluate the mean detection performance. Likewise, we also use CorLoc for PASCAL VOC07/12 and MSCOCO to evaluate the localization performance. More details are shown in Tab. \ref{tab_semi_dataset}.

\begin{table*}[!t]
  \centering
  \caption{Dataset Settings for SS-FSOD. Note that digits represent how many exemplars are labeled for each class. }
    \begin{tabular}{|c|c|}
    \hline
    Methods & Detail \\
    \hline
    MSPLD \cite{dong2017few} & PASCAL VOC07/12 (CorLoc/mAP, 3), MSCOCO (CorLoc/mAP, 3), ImageNet-LOC (CorLoc/mAP, 3) \\
    \hline
    DETReg \cite{bar2021detreg} & MSCOCO (mAP/mAP75, 10/30)\\
    \hline
    CGDP \cite{li2021few}& PASCAL VOC07/12 (mAP, 1/2/3/5/10), MSCOCO (mAP/mAP75, 10) \\
    \hline
    TIP \cite{li2021transformation} & PASCAL VOC07/12 (mAP, 1/2/3/5/10) \\
    \hline
    \end{tabular}%
    \label{tab_semi_dataset}
\end{table*}%

\subsection{Solutions}
\subsubsection{Self-training Based Methods}
Self-training is a classic way in semi-supervised learning, aiming to extract proper pseudo labels as ground-truth labels for unannotated data. As aforesaid, a classic self-training based approach tends to use all labeled data to pre-train a teacher detector, then applies it to generate pseudo labels for all unlabeled images, and finally samples a set of robust pseudo labels to learn a student detector. The key for self-training based methods is the quality of pseudo labels and it is sub-optimal to exploit a single teacher detector for pre-generated pseudo labels or only update pseudo labels once to train a student detector, which can limit its performance.

\begin{figure*}[!t]
\centering
\includegraphics[width=\textwidth]{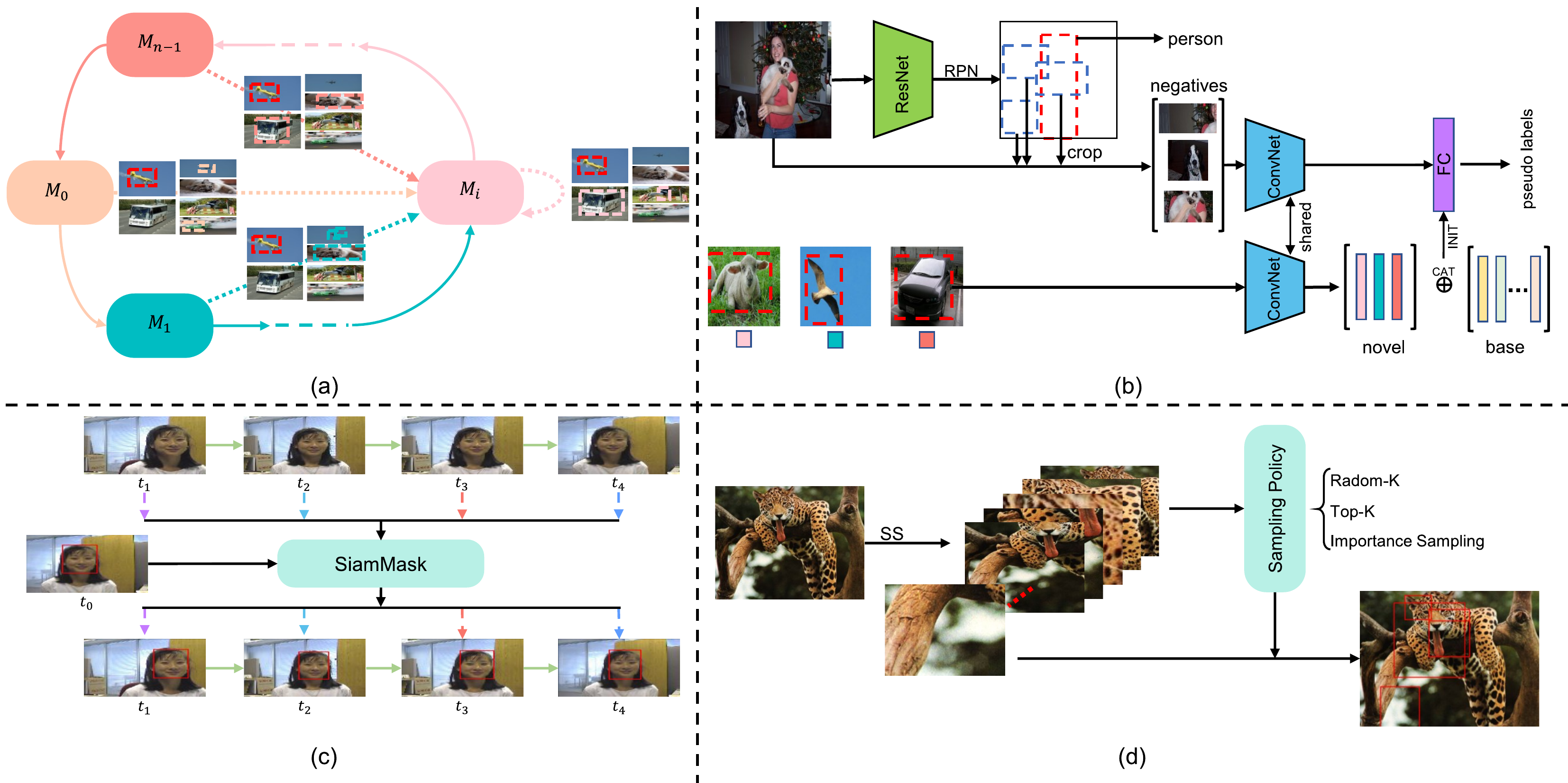}
\caption{The pipeline of four classic pseudo-label generations. (a) MSPLD \cite{dong2017few}; (b) DETReg \cite{bar2021detreg}; (c) FCOS Ensemble++ \cite{yoon2021semi}; (d) CGDP \cite{li2021few}.}
\label{fig_pseudo_labels}
\end{figure*}

\begin{itemize}
\item{\textbf{MSPLD. } Dong et al. \cite{dong2017few} proposed MSPLD that was based on self-learning and multi-modal learning to obtain high-quality pseudo labels for robust training, as illustrated in Fig. \ref{fig_pseudo_labels}(a). It could be concluded as three strategies: (1) \emph{Hard Example Removing.} Dong et al. first assumed that images with over 4 pseudo boxes for a specific class or over 4 classes were likely to have no reliable labels, which should be removed during each iteration step; (2) \emph{Model Ensemble.} Multiple detectors were utilized together to produce raw labels, and NMS with confidence-based box filtering was applied on raw labels for pseudo ground-truth labels; (3) \emph{Training Pool.} It was defined as an indicator whether or not an image should be used in training phase. Specially, it would affect an image’s access threshold to training pools for a specific model, when this model had a number of images used at the last iteration, low discrepancies among proposals and pseudo labels and high consistency exploiting such an image in other detectors’ training pools. In training stage, Dong et al \cite{dong2017few} adopted an iterative policy to update pseudo labels, training pool and parameters of a specific model one by one, i.e., solid lines in Fig. \ref{fig_pseudo_labels}(a).}

\item{\textbf{DETReg.} Bar et al. \cite{bar2021detreg} adopted a two-step training pipeline which disentangled pseudo labels and ground-truth labels for training, avoiding to bring too much noise into the fine-tuning stage, as depicted in Fig. \ref{fig_pseudo_labels}(b). Thus, pseudo labels were exploited in a pretext task to learn how to localize objects and how to encode robust features.

In such a pretext task, it mainly relied on a hypothesis that, compared with non-object boxes, object boxes should have less variations where DERT should learn not only to distinguish such variations but also to cluster automatically. To tackle it, several prediction heads were added upon DERT, i.e., a bounding box regressor for localization, a feature encoder for SwAV descriptor matching and a foreground-background classifier. As for pseudo labels, Bar et al. exploited SS to produce thousands of RoIs for high recall. In view of low precision for SS, a list of sampling strategies (i.e., Top-K, Random-K and Importance Sampling) were presented to fetch proper RoIs as ground-truth boxes (labeled by 1). Specially, Top-K showed uniformly optimal results in the testing phase, due to its relatively robust pseudo labels. In fine-tuning phase, we only saved the bounding box regressor while this binary classifier was replaced with a multi-category classifier to fit a specific task.}

\item{\textbf{CGDP.} In limited-supervised FSOD, it usually split existing benchmarks (e.g., Pascal VOC07/12) into a large-scale base dataset and a small-scale novel dataset, as illustrated in Fig. \ref{fig_pseudo_labels}(d). It was inevitable to contain novel/base objects in the base/novel dataset which was bad for the few-shot learner (equivalent to incomplete labeling). Actually, it could be treated as a semi-supervised problem where unlabeled regions in an image should be handled carefully. Li et al. \cite{li2021few} proposed a similarity-based strategy to inspect negative proposals, which had low IoU with all ground-truth annotations, for novel instance mining. First, weight imprinting was applied to extract a normalized intra-class mean prototype per novel class. Then, only high-possibility negative proposals were sampled to match various class prototypes by cosine similarity for pseudo-label generation.}
\end{itemize}

\subsubsection{Self-Supervised Based Methods}
Although pseudo labels could provide guidance information, it also brings noise into training student detectors. Thus, it is necessary to exploit other inherent relations among unannotated data. Apart from localizing and classifying as accurately as possible, a qualified detector should meet several requirements, e.g., robust feature encoding. Naturally, we could simply exploit unlabeled images to enforce backbones to encode discriminative features which could be applied to recover raw image signals as possible. In the semi-supervised learning, several works have developed associated pretext tasks, such as VAE/GAN based reconstruction, \cite{zhao2015stacked,maaloe2016auxiliary,odena2016semi,dai2017good}. In 
some way, consistency regularization can be regarded as a supervised solution to extract underlying information from unlabeled data.
\begin{itemize}
\item{\textbf{TIP.} In low-shot scenarios, data distributions appear discontinuous very commonly, and naive data augmentation could make it worse which could lower inter-category distance. Especially, a metric-based approach tends to use an encoder to generate category prototypes which is crucial for category matching. In this case, an encoder should be invariant to any kinds of transformations. Thus, Li et al. \cite{li2021transformation} utilized unlabeled images to add a consistent constraint for transformation invariance where features of both an image and its corrupted image should be consistent/similar.}
\end{itemize}

\section{Weakly-Supervised Few-Shot Object Detection}
\label{sec_weakly}
WSOD tends to require 10/50 percent of images with manual tags in datasets, e.g., Pascal VOC07/12 \cite{everingham2010pascal} and MSCOCO \cite{chen2015microsoft}. Compared with instance-level boxes, tags save unit labor cost while it fails to reduce the size of a dataset needing to be labeled. It still brings large annotation burdens into data preparation. Thus, an intuitive solution is to exploit a few images with tags for each category in training phase (usually no more than 200). We properly relax restraints on the size of dataset with image-level labels due to its imprecise signals. Here, we will first give a more formal definition for WS-FSOD.

\subsection{Problem Definition}
As aforesaid, $D_{novel}^+$ denotes a small dataset with image-level labels or object locations and $D_{novel}^-$ denotes an optional large unlabeled dataset sampled in the target domain. For each sample $(I,Y)$ in $D_{novel}^-$, $I$ is an image ($I\in\mathbb{R}^{M\times N\times3}$) and $Y=\{y_n\}^N$ is a subset of $C_{novel}$ which appears in $I$. In some cases, we also have a dataset without target supervision to learn generic notions instead of training from scratch.
Let $D_{novel}^n$ be a set of images which contains instances of the n-th category in $D_{novel}$. Like in LS-FSOD, we restrict the maximum of the number of images per category in $D_{novel}$: $\max{\{|D_{novel}^i|,\ i\in C_{novel}\}}\le k$ (k is usually no more than 200). In all, WS-FSOD aims at actively associating image-level tags with all object regions to learn a robust learner which could be competent to the fully-supervised counterpart.

\subsection{Dataset}
Here, we list dataset settings of all existing WS-FSOD approaches, as shown in Tab. \ref{tab_weakly_dataset}. Existing approaches still exploit several benchmarks to build a weakly-supervised few-shot benchmark, including KITTI, Cityscapes, VisDA-18, MSCOCO, PASCAL VOC 07/12, ImageNet-LOC and CUB. In these excellent benchmarks, they could be grouped into two types, i.e., intra-domain and cross-domain benchmarks. Su et al. \cite{su2020active} combined two distinct benchmarks (i.e., KITTI and Cityscapes) to construct a cross-domain benchmark. Another common practice is to split a benchmark into two disjoint sets to form a intra-domain benchmark.
\begin{table*}[!t]
  \centering
  \caption{Dataset Settings for WS-FSOD. Note that digits represent how many exemplars are labeled for each class. }
    \begin{tabular}{|c|c|}
    \hline
    Methods & Detail \\
    \hline
    AADA \cite{su2020active} & KITTI $\rightarrow$ Cityscapes (mAP, 10/20/30/50/100/200), VisDA-18 (mAP, N/S) \\
    \hline
    vMF-MIL \cite{shaban2021few} & MSCOCO (mAP, 5/10), PASCAL VOC 07 (mAP, 5/10) \\
    \hline
    StarNet \cite{karlinsky2020starnet} & ImageNet-LOC (mAP30/mAP, 1/5), CUB (mAP30/mAP, 1/5), PASCAL VOC (mAP30/mAP, 1/5) \\
    \hline
    NSOD \cite{yang2021training} & PASCAL VOC07/12 (mAP, 20) \\
    \hline
    \end{tabular}%
    \label{tab_weakly_dataset}
\end{table*}%

\subsection{Solutions}
As WS-FSOD mainly attempts to exploit a few image-level labels, its approaches need not only to tackle classic issues in object detection, such as occlusion, deformation, domain shifts, various imbalance problems and so on, but also to confront with imprecise and less annotations. Owing to less supervisory signals, generic WSOD approaches cannot figure out underlying relations among image-level tags and RoIs, where it further increases data uncertainty instead. More specifically, low sampling density easily leads to large intra-category variations or low inter-category distance and such loose data streams can not only have almost no statistic laws but also hinder normal label propagation in the learning processes of WSOD. It is crucial to exploit weakly-labeled data more efficiently and eliminate uncertainty as much as possible.

So far, WS-FSOD has not received enough attention in the computer vision community. Here, it mainly includes three types, i.e., active adversarial learning, multiple instance learning and metric-based learning.
\begin{itemize}
\item{\textbf{AADA.} Su et al. \cite{su2020active} employed adversarial learning to force feature extractors to transform a WS-FSOD problem into a fully-supervised problem with a large-scale dataset for full use of source-domain knowledge. If a specific image cannot fit source domain well, detectors would apply non-matched notions to get poor results. Thus, Su et al. \cite{su2020active} also applied a well-optimized discriminator to indicate whether unlabeled data was similar to labeled data or not. A subset of exemplars with high distinction was sampled to obtain image-level tags to mitigate outliers which cannot fit source domain well. What’s more, Su et al. \cite{su2020active} explored several sampling strategies to select images for maximizing performance gain, i.e., importance weight, K-means clustering, K-center, diversity, Best-versus-Second Best and random selection.}
\item{\textbf{vMF-MIL}. Shaban et al. \cite{shaban2021few} proposed vMF-MIL that was a probabilistic multiple instance learning approach for WS-FSOD. It employed a two-step strategy. It first obtained a base learner from a large-scale dataset with instance-label boxes. To utilize source-domain notions as much as possible, Shaban et al. \cite{shaban2021few} reformed Faster R-CNN \cite{ren2015faster} and make it fully category-agnostic by removing its traditional multi-category classifier. During the pseudo-label generation, it held a statistic assumption that RoI features for each category should form a cluster and have high distinction with that for other categories. For a specific category, it applied EM clustering on those features for finding exact a pseudo label with highest confidence for each image whose tags should include such a category. Then, an off-the-shelf FSOD approach \cite{wang2020frustratingly} could be applied to learn a novel task with pseudo boxes.}

\item{\textbf{StarNet.} Karlinsky \cite{karlinsky2020starnet} proposed StarNet that was also a metric-based solution. It mainly exploited feature co-occurrences for reliable evidence to mine all instances in query images without pseudo-label generation. It also consisted of two important fusion nodes: (1) Voting Heatmaps. It first measures a point-to-point similarity map by L1 norm, and calculates all permutations as voting heatmaps for matching support and query features. (2) Back-Projection Maps. It takes a permutation with highest matching score to suppress low-quality evidence. Except for an image-level category constraint on query features, it performed a consistency constraint among support and query features, which could form a close cluster for each category as well.}
\item{\textbf{NSOD.} Yang et al. \cite{yang2021training} proposed NSOD that was a metric-based MIL-style framework which attempted to mine implicit relations among intra-category mean prototypes and all RoIs. Especially, it could directly start from a small set of annotated images to propagate image-level tags to instance-level boxes for unlabeled data. It first utilized a backbone pretrained in ImageNet to extract features for both support images and all RoIs of query images (without any supervision). Then, for a specific category, all global features were averaged as an intra-category mean prototype which was combined with all RoI features to obtain pseudo labels for each query image. Next, pseudo labels were exploited to learn a teacher which was exploited to refine pseudo labels. Finally, two kinds of pseudo labels are averaged to form ground-truth pseudo labels for student learning. It was inevitable to include noisy background information into these intra-category mean prototypes, which was detrimental to final results. }
\end{itemize}

\section{Conclusion}
\label{sec_conclusion}
Few-shot detectors have obtained some key achievements to mitigate urgent need of abundant labeled training data for classic deep-learning architectures, playing an important role in various applications, such as wise medical \cite{quellec2020automatic, wei85few}. Therefore, we provide a comprehensive survey on few-shot learning for object detection. To provide a detailed analysis of this few-shot issue, this survey introduces a data-based taxonomy according to training data and associated supervisory signals of a novel task in terms of definitions, datasets, criteria, strengths and weaknesses for each kind of approaches. Besides, we discuss main challenges that needed to be tackled and how these approaches interplay and boost performance. 

As aforesaid, existing few-shot detectors still need to tackle some imperative problems (e.g., domain shifts) and have a huge performance gap with many-shot detectors and human. Here, we will discuss future trends in this promising domain.

\textbf{Domain Transfer.} In a real-world task, it is not always easy to find a suitable dataset that could be exploited for cross-domain knowledge and have low domain shifts with this task. Although many works attempted to employ GAN variants to align the semantic space between the source and target domain, task-specific notions could be non-matched with the source domain and easily lost in this alignment process due to lack of associated supervision, especially with large domain shifts \cite{sankaranarayanan2018generate, chen2018semantic,chen2020multiple}. As active-learning strategies in \cite{su2020active}, we notice that it is key to combine effectively and efficiently incorporate cross-domain and task-specific knowledge for better domain transfer.

\textbf{Efficient Architectures.} Existing solutions for few-shot object detection still inherit the architectures designed for generic object detection \cite{girshick2014rich,ren2015faster,redmon2017yolo9000,liu2016ssd,redmon2018yolov3,bochkovskiy2020yolov4} which contain abundant learnable parameters. In these common architectures, more data and layers (parameters) usually mean higher performance. Thus, a novel and efficient architecture should be designed to tackle this issue while there is no work in this direction.

\textbf{Robust Feature Extractors.} In general, feature extractors are viewed as a task-agnostic component, which is typically pre-trained in a large dataset and kept fixed in another task. As in \cite{singh2018sniper}, feature extractor is susceptible to scale variations. Meanwhile, existing architectures extremely rely on the features produced by backbones for dense proposals generation, which is crucial to the final performance. Thus, there has been growing interest in learning a robust feature extractor \cite{li2021transformation,wu2020multi,chen2018lstd}.

\textbf{Mixed-Supervised Learning.} In WS-FSOD, it is hard to use a few images with image-level tags to associate underlying objects with these image-level tags. If we provide a small training set with instance-level labels except for the aforementioned weakly-labeled data, it only slightly increases annotation costs while it could give a definite signal for low-shot detectors to have more high-quality proposals when compared with weakly-supervised few-shot detectors. However, there are limited works in the mixed-supervised few-shot object detection \cite{chen2019progressive}. According to the above analysis, we think that mixed-supervised few-shot object detection may be a hot research direction. 

\textbf{Unsupervised Learning.} Currently, state-of-the-art architectures have an urgent need for abundant labeled training data to learn an excellent detector. However, in some cases, training data is hard to be collected. Even if we cost significantly to collect a large annotated dataset, it could still contain nasty issues, such as foreground-foreground imbalance and data bias, which go against the normal training process. Therefore, it is essential to study a strategy \cite{zhu2007unsupervised,xie2017aggregated} to train CNNs without the need to collect and annotate a large-scale dataset for a novel task.

\textbf{Data Augmentation.} Data augmentation is a considerably intuitive way to alleviate overfitting by increasing intra-category variations. However, large intra-category variations could relatively lower inter-category distinction, which is interferential to explore suitable category prototypes. Zhang et al. \cite{zhang2021hallucination} proposed a meta strategy that moderately increased the intra-category diversity which could be transferred from other large-scale datasets while it cannot work with large domain shifts, since the source-domain intra-category diversity could not fit the target domain. A better choice is to actively measure intra-category variations and inter-category distance to provide suitable strategies for data augmentation.

\bibliographystyle{IEEEtran}
\bibliography{IEEEabrv,ref}

\begin{sidewaystable*}[!t]
\tiny
  \centering
  \caption{Summarization of metric-based methods for the $N$-Way $K$-Shot limited-supervised problem. Abbreviations: Region Proposal Network (RPN), EdgeBoxes (EB) \cite{zitnick2014edge}, Concatenate (CAT), Embedding Network of Support/Query Branch (S/Q). See Tab. \ref{tab_fusion_node} \& Fig. \ref{fig_fusion_nodes} for more details about fusion nodes.}
  \label{metric_learning_all_methods_review}
    \begin{tabularx}{\textwidth}{|c|c|c|c|c|c|X|}
    \hline
    \makecell{Method} & \makecell{Pipeline \\ Used} & \makecell{Data \\ Preprocessing} & \makecell{Embedding \\ Network} & \makecell{Fusion \\ Node} & \makecell{Published \\ In} & \multicolumn{1}{c|}{Full Name of Method and Highlights} \\
    \hline
    AFD-Net  \cite{liu2020afd} & \makecell{Faster \\ R-CNN} & Mask+CAT & \makecell{ResNet-101 \\ ResNet-50} & \makecell{a-5 \\ a-6} & arXiv2021 & \textbf{(Adaptive Fully-Dual Network, AFD);} Propose AFD to generate task-specific vectors for support \& query; Combine multiple aggregators for better feature fusion. \\
    \hline
     FSOD  \cite{fan2020few} & \makecell{Faster \\ R-CNN}  & Crop+Context & ResNet-50 & \makecell{a-2 \\ a-5} & CVPR2020 & Include 16-pixel image context for better class prototypes; Reweight query features by class-specific vectors per class to make RPN generate high-quality proposals (meta RPN); Use global, local \& patch matching to measure similarity more accurately.  \\
    \hline
    CoAE-Net \cite{hsieh2019one} & \makecell{Faster \\ R-CNN} & Crop  & ResNet-50 & \makecell{a-1 \\ a-3 \\ a-5} & NIPS2019 & \textbf{(Co-Attention and Co-Excitation Network, CoAE);} Use a non-local variant to provide aligned class specific features for RPN to make high-quality proposals; Take a shared SE block to learn for further refine query \& support features. \\
    \hline
     \makecell{META \\ RCNN \cite{wu2020meta}}   & \makecell{Faster \\ R-CNN} & \makecell{Crop from \\ Features} & \makecell{VGG-16 \\ ResNet-50} & a-1   & \makecell{ACM \\ MM2020} & Implicitly enhance category prototypes with contextual information outside instances. \\
    \hline
     AIT-Net \cite{chen2021adaptive}& \makecell{Faster \\ R-CNN} & Crop  & ResNet-50 & \makecell{a-1\\ a-4} & CVPR2021 & Introduce multi-head attention to explore cooccurrences between query \& support features; Employ a transformer variant to explore proposal-support attention. \\
    \hline
    ZSD \cite{bansal2018zero} & R-CNN & N/A   & \makecell{Q: Inception \\ S: MLP} & a-5   & ECCV2018 & \textbf{(Zero-Shot Object Detection, ZSD);} Take a MLP to transform word embedding from semantic space to visual space (i.e., cross-domain alignment); Assign a list of background images with a set of categories randomly sampled from corpus, which is disjoint with seen classes, to force ZSD to learn a latent background concept. \\
    \hline
    NP-RepMet \cite{yang2020restoring} & \makecell{Faster \\ R-CNN} & {N/A} & \makecell{ResNet-101 \\ (DCN, FPN)} & {a-5} & {NIPS2020} & \textbf{(Representative-based metric learning with negative and positive information, NP-RepMet);} Employ negatives ($0.2<IoU(P,G)<0.3$) to learn negative prototypes; Take the spectral clustering to get K negative prototypes for inference. \\
    \hline
    \makecell{Meta \\ DETR \cite{zhang2021meta}} & \makecell{Deformable\\ DETR} & Crop  & ResNet-101 & a-1   & arXiv2021 & Adopt a novel BERT to get rid of heuristic design in Faster R-CNN; Leverage skip connection to regularize low-/high-level features. \\
    \hline
    \makecell{S-FRCNN \cite{michaelis2020closing}} & \makecell{Faster \\ R-CNN} & Crop  & \makecell{ResNet-50 \\ ResNetXt-101} & a-1   & arXiv2020 & \textbf{(Siamese Faster R-CNN, S-FRCNN);} Explore SUB for feature fusion. \\
    \hline
     DAnA \cite{chen2021should} & \makecell{Faster \\ R-CNN \\ / \\ RetinaNet} & Crop  & ResNet-50 & \makecell{a-2 \\ a-4} & TMM2021 & \textbf{(Dual-Awareness Attention, DAnA);} BA block first generate a foreground heatmap, then sum up features reweighted by such a heatmap along spatial axes to generate attention vectors, and finally add it to original features; CISA is an affinity matrix for semantic alignment. \\

    \hline
    DCNet \cite{hu2021dense} & \makecell{Faster \\ R-CNN} & Crop  & ResNet-101 & a-1   & CVPR2021 & Design a non-local variant which captures global similarity maps between query \& support features of each category, then perform softmax on such similarity maps among all support categories, and finally use normalized maps to refine query features; Take serval RoI-Align layers to capture features at various scales. \\

    \hline
    OSOD \cite{li2020one} & \makecell{Faster \\ R-CNN} & Crop  & ResNet-50 & \makecell{a-2 \\ a-4} & arXiv2020 & \textbf{(One-Shot Object Detection, OSOD);} Take FPN to get a global prototypes per class which is used to compute a cosine similarity map for RPN at each scale; Take a 1x1 conv to correlate query \& support features at each pixel (spatial semantics); Pair a query image with the most similar support instance to alleviate intra-class variance in the training stage; Sample negative boxes with certain IoU with gt boxes to force a classifier to distinguish low-quality proposals. \\

    \hline
    PNSD \cite{zhang2020few} & \makecell{Faster \\ R-CNN} & Crop  & ResNet-50 & \makecell{a-2 \\ a-4} & ACCV2020 & Propose a multi-scale feature aggregator to enhance features of both query \& support; Adopt SOP/PN to capture feature cooccurrences; Design a SN module to combine global/local/patch matching to evaluate similarity more accurately. \\

    \hline
    FSODM \cite{deng2020few} & YOLOv3 & Crop  & \makecell{Q: DarkNet-53 \\ S: N/S} & b-1   & TGRS2020 & Take a FPN variant to learn features of different scales to alleviate large-scale variations. \\

    \hline
    CME \cite{li2021beyond} & YOLOv2 & Mask+CAT & \makecell{Q: DarkNet-19 \\ S: N/S} & a-4   & CVPR2021 & \textbf{(Class Margin Equilibrium, CME);} Design a max-margin loss to guarantee that intra-class distance is less than inter-class distance as much as possible in meta-training stage; Erase discriminative pixels in a support mask by its gradient map for data augmentation. \\

    \hline
    FSOD \cite{zhang2020few} & \makecell{Faster \\ R-CNN} & Crop  & \makecell{ResNet-50 \\ ResNet-101} & \makecell{a-1 \\ a-5} & CVPR2021 & Average similarity (scores) at all levels by a RN variant to get a reliable score for each RoI; Dynamically assign weights to determine class prototypes by similarity between query \& support instead of simple average to learn a robust detector; Design a kernel generator to dynamically generate support-specific kernels to explore feature cooccurrences; Take a adaptive margin to improve a contrastive loss to distinguish a correct category from background and other foreground categories; Adopt focal loss to tackle the foreground-background imbalance problem. \\

    \hline
    \makecell{Meta Faster \\ R-CNN \cite{han2021meta}} & \makecell{Faster \\ R-CNN} & Crop+Context & \makecell{ResNet-50 \\ ResNet-101} & \makecell{a-2 \\ a-5} & arXiv2021 & Combine MUL, SUB \& CAT to generate more robust features; Calculate an affinity matrix of RoIs and all category prototypes for spatial alignment, and then sum up such an affinity matrix by row for background suppression. \\

    \hline
    TIP \cite{li2021transformation} & \makecell{Faster \\ R-CNN} & Mask+CAT & N/S   & a-4   & CVPR2021 & \textbf{(Transformation Invariant Principle, TIP);} Naive data augmentation may have negative effect on FSOD; Design a consistency regularization to learn a robust backbone to narrow intra-class variations (e.g., rotation, occlusion), while a CE loss is adopted to expand inter-class distance for category prototypes; Design a proposal consistent regularization which takes transformed images to predict RoIs, and then get such RoIs features from features of original images to make final results. \\

    \hline
    MM-FSOD \cite{li2020mm} & \makecell{Faster \\ R-CNN} & Crop  & ResNet-34 & a-5   & arXiv2020 & \textbf{(Meta and Metric integrated Few-Shot Object Detection, MM-FSOD);} Take Pearson distance to distinguish hard negative proposals; View background patches as instances of the background category. \\

    \hline
    ONCE \cite{perez2020incremental} & CenterNet & N/S   & ResNet-50 & b-1   & CVPR2020 & \textbf{(OpeNended Centre nEt, ONCE);} Fix all category-agnostic parameters in the finetuning stage; Learn a generator to generate a category representation for each novel category. \\

    \hline
    FSOD & YOLOv2 & Mask+CAT & \makecell{Q: DarkNet-19 \\ S: N/S} & b-1   & ICCV2019 & Design a feature reweighting module to generate category-specific features by a given category to detect all instances of such a category. \\

    \hline
    RepMet \cite{karlinsky2019repmet} & \makecell{Faster \\ R-CNN} & N/A   & Inception & a-5   & CVPR2019 & \textbf{(Representative-based metric learning, RepMet);} Take K category prototypes for each category to match each RoI; Design an embedding loss (a kind of margin loss) to minimize intra-category variations and maximize inter-category distance. \\

    \hline
    \makecell{Meta \\ R-CNN \cite{yan2019meta}} & \makecell{Faster \\ R-CNN} & Mask+CAT & ResNet-101 & a-4   & ICCV2019 & Propose a meta loss to make category prototypes distinguish each other. \\

    \hline
    SRR-FSD \cite{zhu2021semantic} & \makecell{Faster \\ R-CNN} & N/A   & \makecell{Q:ResNet-101 \\ S:A knowledge graph} & a-5   & CVPR2021 & (Semantic Relation Reasoning for Shot-Stable Few-Shot Object Detection, SRR-FSD); Propose a knowledge graph to model inter-category relations; Add fc layers respectively to provide more suitable features for the classifier and locator; Design a MLP for cross-domain alignment.  \\

    \hline
    \end{tabularx}%
\end{sidewaystable*}%

\begin{sidewaystable*}[!t]
\tiny
  \centering
  \caption{Summarization of transfer-learning methods for the N-Way K-Shot limited-supervised problem.}
    \begin{tabularx}{\linewidth}{|c|c|c|c|X|}
    \hline
    Method & \makecell{Pipeline \\ Used} & Backbone & \makecell{Published \\ In} & \multicolumn{1}{c|}{Full Name of Method and Highlights} \\
    \hline
    CT-Net \cite{yang2020context} & SSD   & VGG-16 & AAAI2020 & \textbf{(Context-Transformer Network, CT-Net);} Adopt max-pooling to capture more contextual information outside a prior anchor; Generate an affinity matrix between features of prior anchor w/ \& w/o contextual fields to refine features of prior anchor for more semantic context to classify each proposal. \\
    \hline
    LEAST \cite{li2021class}& \makecell{Faster \\ R-CNN} & ResNet-101 & arXiv2021 & \textbf{(Less forgetting, fEwer training resources, And Stronger Transfer capability, LEAST);} Make a trade-off between two kinds of updating rules (i.e., FIX-ALL \& FIT-ALL) to fine-tune parameters of the Faster R-CNN head; Employ KL divergence to make consistent probability distribution of base categories in the fine-tuning phase; Take a cluster-based exemplar selection algorithm to choose the most representative K exemplars of each base category instead of randomly sampling. \\
    \hline
    \makecell{Retentive \\ R-CNN \cite{fan2021generalized}} & \makecell{Faster \\ R-CNN} & ResNet-101 & CVPR2021 & Adopt a cosine classifier to filter drastic variations of feature norms between base \& novel categories; Take KL divergence to maintain probability distribution of base categories in the fine-tuning stage; Add an extra category-agnostic classifier for novel categories to capture more high-quality proposals of novel categories as possible; Assign more weight on a base learner to predict a proposal of base categories to make more reliable results in inference. \\
    \hline
    $FSOD^{up}$ \cite{wu2021universal} & \makecell{Faster \\ R-CNN} & ResNet-101 & arXiv2021 & Learn a set of category-agnostic prototypes (e.g., shared features) to enhance features of a given image to make more high-quality proposals; Apply a learnable affine transformation on universal prototypes to generate conditional prototypes which are employed to provide discriminative features for the further locator and classifier; Employ a consistent constraint (KL divergence) between conditional \& augmented probability distributions produced by a shared classifier which takes conditional \& augmented prototypes as inputs to learn a more robust detector. \\
    \hline
    AttFDNet \cite{chen2020leveraging} & RFB Net & VGG-16 & arXiv2020 & \textbf{(Attentive Few-Shot Object Detection Network, AttFDNet);} Employ a saliency sub-network to force AttFDNet to pay more attention to foreground regions; Combine a pixel-wise convolution layer with a softmax layer to generate a heatmap, and sum up such a heatmap along spatial axes to generate global representations for semantic alignment; Leverage RFB block to capture more context information without excessive parameters; Design a object-concentration loss to maximize cosine similarity between all positive proposals \& associated category-specific weights; Leverage OHEM to enforce AttFDNet to distinguish low-quality positive proposals. \\
    \hline
    FAFRCNN \cite{wang2019few} & \makecell{Faster \\ R-CNN} & VGG-16 & CVPR2019 & \textbf{(Few-shot Adaptive Faster-RCNN, FAFRCNN);} Design split pooling to randomly crop feature patches by a set of predefined anchors from source \& target domain to form source-source \& source-target pairs for domain-adversarial learning; Randomly sample RoIs from source \& target domain to semantically align paired object features; Take L2 penalty to make consistent features between the pre-trained network and the fine-tuned network to alleviate overfitting. \\
    \hline
    LSTD \cite{chen2018lstd}& \makecell{Faster \\ R-CNN} & VGG-16 & AAAI2018 & \textbf{(Low-Shot Transfer Detector, LSTD);} Design a BD regularization which take ground-truth boxes of a given image to generate a background mask to depress background regions; Propose a KL divergence to utilize cross-domain similarity to transfer general knowledge. \\
    \hline
    TFA \cite{wang2020frustratingly} & \makecell{Faster \\ R-CNN} & ResNet-101 & arXiv2020 & \textbf{(Two-stage Fine-tuning Approach, TFA);} Design a two-stage training approach which first trains a base learner on a large-scale base benchmark, and then only fine-tunes Faster R-CNN head on a small-scale novel benchmark; Take a cosine classifier to reduce intra-category variance; Revise evaluation protocols that use more repeated runs to reduce sample variance for fair comparison. \\
    \hline
    MPSR \cite{wu2020multi} & \makecell{Faster \\ R-CNN} & ResNet-101 & ECCV2020 & \textbf{(Multi-Scale Positive Sample Refinement, MPSR);} Randomly sample a batch of instances in an image, and resize such instances to various scales/aspect ratios (object pyramid) for more positive training data to refine FPN. \\
    \hline
    FSCE \cite{sun2021fsce} & \makecell{Faster \\ R-CNN} & ResNet-101 & CVPR2021 & \textbf{(Few-Shot Object Detection via Contrastive Proposals Encoding, FSCE);} Reuse positive proposals with low confidence suppressed by NMS \& remove negative proposals by half to rebalance gradient; Design a contrastive proposal encoding (CPE) loss to learn a discriminative feature extractor. \\
    \hline
    \end{tabularx}%
    \label{details_transfer_learning}
\end{sidewaystable*}%

\end{document}